\PassOptionsToPackage{unicode}{hyperref}
\PassOptionsToPackage{hyphens}{url}
\PassOptionsToPackage{dvipsnames,svgnames,x11names}{xcolor}
\documentclass[
]{interact}

\usepackage{amsmath,amssymb}
\usepackage{lmodern}
\usepackage{iftex}
\ifPDFTeX
  \usepackage[T1]{fontenc}
  \usepackage[utf8]{inputenc}
  \usepackage{textcomp} % provide euro and other symbols
\else % if luatex or xetex
  \usepackage{unicode-math}
  \defaultfontfeatures{Scale=MatchLowercase}
  \defaultfontfeatures[\rmfamily]{Ligatures=TeX,Scale=1}
\fi
\IfFileExists{upquote.sty}{\usepackage{upquote}}{}
\IfFileExists{microtype.sty}{% use microtype if available
  \usepackage[]{microtype}
  \UseMicrotypeSet[protrusion]{basicmath} % disable protrusion for tt fonts
}{}
\makeatletter
\@ifundefined{KOMAClassName}{% if non-KOMA class
  \IfFileExists{parskip.sty}{%
    \usepackage{parskip}
  }{% else
    \setlength{\parindent}{0pt}
    \setlength{\parskip}{6pt plus 2pt minus 1pt}}
}{% if KOMA class
  \KOMAoptions{parskip=half}}
\makeatother
\usepackage{xcolor}
\ifx\paragraph\undefined\else
  \let\oldparagraph\paragraph
  \renewcommand{\paragraph}[1]{\oldparagraph{#1}\mbox{}}
\fi
\ifx\subparagraph\undefined\else
  \let\oldsubparagraph\subparagraph
  \renewcommand{\subparagraph}[1]{\oldsubparagraph{#1}\mbox{}}
\fi

\usepackage{longtable,booktabs,array}
\usepackage{calc} % for calculating minipage widths
\usepackage{etoolbox}
\makeatletter
\patchcmd\longtable{\par}{\if@noskipsec\mbox{}\fi\par}{}{}
\makeatother
\IfFileExists{footnotehyper.sty}{\usepackage{footnotehyper}}{\usepackage{footnote}}
\makesavenoteenv{longtable}
\usepackage{graphicx}
\makeatletter
\def\maxwidth{\ifdim\Gin@nat@width>\linewidth\linewidth\else\Gin@nat@width\fi}
\def\maxheight{\ifdim\Gin@nat@height>\textheight\textheight\else\Gin@nat@height\fi}
\makeatother
\setkeys{Gin}{width=\maxwidth,height=\maxheight,keepaspectratio}
\makeatletter
\def\fps@figure{htbp}
\makeatother
\newlength{\cslhangindent}
\newlength{\csllabelwidth}
\newlength{\cslentryspacingunit} % times entry-spacing
\newenvironment{CSLReferences}[2] % #1 hanging-ident, #2 entry spacing
 {% don't indent paragraphs
  \setlength{\parindent}{0pt}
  \ifodd #1
  \let\oldpar\par
  \def\par{\hangindent=\cslhangindent\oldpar}
  \fi
  \setlength{\parskip}{#2\cslentryspacingunit}
 }%
 {}
\usepackage{calc}

\usepackage{booktabs}
\usepackage{longtable}
\usepackage{array}
\usepackage{multirow}
\usepackage{wrapfig}
\usepackage{float}
\usepackage{colortbl}
\usepackage{pdflscape}
\usepackage{tabu}
\usepackage{threeparttable}
\usepackage{threeparttablex}
\usepackage[normalem]{ulem}
\usepackage{makecell}
\usepackage{xcolor}
\usepackage{orcidlink}
\usepackage{graphicx}
\usepackage{hyperref}
\usepackage[utf8]{inputenc}

\usepackage{setspace}
\usepackage{soul}
\makeatletter
\@ifpackageloaded{caption}{}{\usepackage{caption}}
\AtBeginDocument{%
\ifdefined\contentsname
  \renewcommand*\contentsname{Table of contents}
\else
  \newcommand\contentsname{Table of contents}
\fi
\ifdefined\listfigurename
  \renewcommand*\listfigurename{List of Figures}
\else
  \newcommand\listfigurename{List of Figures}
\fi
\ifdefined\listtablename
  \renewcommand*\listtablename{List of Tables}
\else
  \newcommand\listtablename{List of Tables}
\fi
\ifdefined\figurename
  \renewcommand*\figurename{Figure}
\else
  \newcommand\figurename{Figure}
\fi
\ifdefined\tablename
  \renewcommand*\tablename{Table}
\else
  \newcommand\tablename{Table}
\fi
}
\@ifpackageloaded{float}{}{\usepackage{float}}
\floatstyle{ruled}
\@ifundefined{c@chapter}{\newfloat{codelisting}{h}{lop}}{\newfloat{codelisting}{h}{lop}[chapter]}
\floatname{codelisting}{Listing}

\makeatother
\makeatletter
\@ifpackageloaded{caption}{}{\usepackage{caption}}
\@ifpackageloaded{subcaption}{}{\usepackage{subcaption}}
\makeatother
\makeatletter
\@ifpackageloaded{tcolorbox}{}{\usepackage[many]{tcolorbox}}
\makeatother
\makeatletter
\@ifundefined{shadecolor}{\definecolor{shadecolor}{rgb}{.97, .97, .97}}
\makeatother
\ifLuaTeX
  \usepackage{selnolig}  % disable illegal ligatures
\fi
\IfFileExists{bookmark.sty}{\usepackage{bookmark}}{\usepackage{hyperref}}
\IfFileExists{xurl.sty}{\usepackage{xurl}}{} % add URL line breaks if available
\hypersetup{
  pdftitle={Classification and mapping of low-statured shrubland cover types in post-agricultural landscapes of the US Northeast},
  pdfauthor={Michael J Mahoney; Lucas K Johnson; Abigail Guinan; Colin M Beier},
  pdfkeywords={LiDAR, Landsat, shrubland, machine learning, neural
networks, land cover},
  colorlinks=true,
  linkcolor={blue},
  filecolor={Maroon},
  citecolor={Blue},
  urlcolor={Blue},
  pdfcreator={LaTeX via pandoc}}

\title{Classification and mapping of low-statured shrubland cover types
in post-agricultural landscapes of the US Northeast\thanks{Please cite
as: Mahoney, M. J., Johnson, J. K., Guinan, A. Z., and Beier, C. M.
2022. Classification and mapping of low‑statured 'shrubland' cover types
in post‑agricultural landscapes of the US Northeast. The International
Journal of Remote Sensing, 43(19‑24), 7117‑7138.
https://doi.org/10.1080/01431161.2022.2155086}}
\author{Michael J
Mahoney$\textsuperscript{1}$~\orcidlink{0000-0003-2402-304X}, Lucas K
Johnson$\textsuperscript{1}$~\orcidlink{0000-0002-7953-0260}, Abigail
Guinan$\textsuperscript{1}$~\orcidlink{0000-0002-8860-3089}, Colin M
Beier$\textsuperscript{2}$~\orcidlink{0000-0003-2692-7296}}

\thanks{CONTACT: Michael J
Mahoney. Email: \href{mailto:mjmahone@esf.edu}{\nolinkurl{mjmahone@esf.edu}}. Lucas
K
Johnson. Email: \href{mailto:ljohns11@esf.edu}{\nolinkurl{ljohns11@esf.edu}}. Abigail
Guinan. Email: \href{mailto:aguinan@esf.edu}{\nolinkurl{aguinan@esf.edu}}. Colin
M
Beier. Email: \href{mailto:cbeier@esf.edu}{\nolinkurl{cbeier@esf.edu}}. }
\begin{document}
\maketitle
\textsuperscript{1} Graduate Program in Environmental Science, State
University of New York College of Environmental Science and
Forestry, Syracuse, NY, USA\\ \textsuperscript{2} Department of
Sustainable Resources Management, State University of New York College
of Environmental Science and Forestry, Syracuse, NY, USA
\begin{abstract}
Novel plant communities reshape landscapes and pose challenges for land
cover classification and mapping that can constrain research and
stewardship efforts. In the US Northeast, emergence of low-statured
woody vegetation, or shrublands, instead of secondary forests in
post-agricultural landscapes is well-documented by field studies, but
poorly understood from a landscape perspective, which limits the ability
to systematically study and manage these lands. To address gaps in
classification/mapping of low-statured cover types where they have been
historically rare, we developed models to predict shrubland
distributions at 30m resolution across New York State (NYS), using a
stacked ensemble combining a random forest, gradient boosting machine,
and artificial neural network to integrate remote sensing of structural
(airborne LIDAR) and optical (satellite imagery) properties of
vegetation cover. We first classified a 1m canopy height model (CHM),
derived from a patchwork of available LIDAR coverages, to define
shrubland presence/absence. Next, these non-contiguous maps were used to
train a model ensemble based on temporally-segmented imagery to predict
shrubland probability for the entire study landscape (NYS).
Approximately 2.5\% of the CHM coverage area was classified as
shrubland. Models using Landsat predictors trained on the classified CHM
were effective at identifying shrubland (test set AUC=0.893, real-world
AUC=0.904), in discriminating between shrub/young forest and other cover
classes, and produced qualitatively sensible maps, even when extending
beyond the original training data. After ground-truthing, we expect
these shrubland maps and models will have many research and stewardship
applications including wildlife conservation, invasive species
mitigation and natural climate solutions. Our results suggest that
incorporation of airborne LiDAR, even from a discontinuous patchwork of
coverages, can improve land cover classification of historically rare
but increasingly prevalent shrubland habitats across broader areas.
\end{abstract}
\begin{keywords}
\def\sep{;\ }
LiDAR\sep Landsat\sep shrubland\sep machine learning\sep neural
networks\sep 
land cover
\end{keywords}
\ifdefined\Shaded\renewenvironment{Shaded}{\begin{tcolorbox}[breakable, frame hidden, boxrule=0pt, sharp corners, enhanced, interior hidden, borderline west={3pt}{0pt}{shadecolor}]}{\end{tcolorbox}}\fi

\hypertarget{introduction}{%
\section{Introduction}\label{introduction}}

Human land use has fundamentally altered vegetation-environment
relationships and created legacies that include the emergence of novel
communities and ecosystem types (Foster, Motzkin, and Slater 1998;
Cramer, Hobbs, and Standish 2008). In post-agricultural landscapes of
eastern North America, these legacies include loss of plant diversity
(Flinn and Vellend 2005) and widespread homogenization of vegetation
composition and structure, relative to historical reconstructions
(Foster, Motzkin, and Slater 1998; Flinn, Vellend, and Marks 2005).
Widespread abandonment of crop, pasture and industrial lands from the
late-19th to middle-20th centuries created an expanding land base for
invasion and emergence of novel communities (Williams and Jackson 2007;
Fridley 2012; Alexander, Diaz, and Levine 2015), with variable outcomes
depending on prior land use practices (Stover and Marks 1998; Benjamin,
Domon, and Bouchard 2005; Kulmatiski, Beard, and Stark 2006). Entirely
novel communities have emerged in old-fields due to colonization by
non-native plants, including invasive woody shrubs, which are much more
likely to establish and become dominant in post-agricultural (Johnson et
al. 2006; Cramer, Hobbs, and Standish 2008; McCay and McCay 2009) and
post-industrial sites (Spiering 2019) compared to closed-canopy forests
of any successional age. Meanwhile, secondary forests across the US
Northeast, including those established in old fields, often lack
sufficient advance regeneration to maintain productivity and resilience
to changing disturbance regimes (Dey et al. 2019).

Among the outcomes of these changes, the emergence of low-statured
vegetation or shrublands as a more common cover type in the US Northeast
has been suggested by numerous field studies, but is poorly understood
from a landscape perspective. Here the term shrubland reflects a
physiognomic definition following King and Schlossberg (2014), which
encompasses several types of plant communities found in the US
Northeast, including: 1) young, regenerating or otherwise low-statured
closed-canopy forests; 2) wetlands dominated by native shrubs (e.g.,
\emph{Alnus} spp) or small-statured trees (e.g, \emph{Picea marianas} in
boreal peatlands); 3) uplands dominated by native shrubs (e.g.,
\emph{Cornus racemosa}); and most recently, 4) upland shrub/scrub
dominated by invasive woody (e.g., \emph{Rhamnus} and \emph{Lonicera}
spp) and herbaceous (e.g., \emph{Solidago} spp) perennials. Among the
types above, regenerating forests and invasive shrub/scrub communities
are of growing interest for research and management purposes, given
their anthropogenic origins, their potential novelty in terms of
composition and dynamics, and their implications for biodiversity and
ecosystem services (Cramer, Hobbs, and Standish 2008; Hobbs, Higgs, and
Harris 2009; Perring, Standish, and Hobbs 2013; Dey et al. 2019).
Despite their conservation value as wildlife habitat, especially for
songbirds, shrublands are widely unpopular cover types in terms of their
perceived aesthetic, recreational and economic values (Askins 2001; King
and Schlossberg 2014). Although long disregarded, these lands are
rapidly gaining attention in today's urgent push to implement natural
climate solutions (Fargione et al. 2018) and identify marginal or
underutilized lands for renewable energy generation.

However, current limitations to the classification and mapping of these
cover types pose obstacles to advancing both science and stewardship
opportunities (Hobbs, Higgs, and Harris 2009). Shrublands are a very
challenging cover class to identify from imagery alone, given the
breadth of community types included (as noted above) and the high
variability in density and canopy cover that exists within and among
those community types (King and Schlossberg 2014). In practical terms
this means that, when relying solely on imagery, shrublands encompass a
full gradient from resembling herbaceous or barren land to resembling
closed-canopy conditions (Brown et al. 2020). Furthermore, shrublands
are relatively rare in the US Northeast, making them particularly
challenging to identify with standard classification methods (Bogner et
al. 2018; Haibo He and Garcia 2009). As a result, imagery-based
approaches tend to classify shrubland categories with substantially
lower accuracy than other land cover classes (Wickham et al. 2021; Brown
et al. 2020).

A solution for this problem might be to incorporate additional,
non-imagery sources of remote sensing data into land cover
classification methodologies. LiDAR data collected through airborne
laser scanning can provide essential information for identifying
low-statured vegetation such as early-successional forests (Falkowski et
al. 2009). In combination with imagery, LiDAR data can enable
continuous, broad-scale estimation of canopy heights and other
structural traits which may greatly simplify the task of distinguishing
between low-statured and taller closed-canopy cover types (Ruiz et al.
2018). Unfortunately, the cost and logistical challenges of airborne
LiDAR collection have constrained its availability to smaller extents
and with much longer return intervals than provided by satellite
imagery. Yet if canopy structural estimates from airborne LiDAR could be
used to label a training dataset in order to fit models using satellite
imagery, it should be possible to produce models capable of identifying
shrubland with greater accuracy than those trained on imagery alone,
while being able to map/model a larger and more contiguous spatial
extent than models relying on airborne LiDAR data as predictors. Similar
methods have been used to automate labeling imagery used to train models
for tree and building detection (Zarea and Mohammadzadeh 2016; Huang et
al. 2019), but to our knowledge this data fusion approach has not yet
been applied for shrubland mapping.

Here we explored this approach to map shrubland across New York State
(NYS), a largely post-agricultural landscape containing extensive
old-fields populated mostly by secondary forests or invasive shrub/scrub
communities, which are difficult to parse based on imagery alone. We
leveraged a non-contiguous patchwork of existing large-footprint LIDAR
data sets by creating very high resolution (1 m) canopy height models
that covered approximately 60\% of the study area (NYS). By sampling
from these canopy height maps, we trained machine learning models with
temporally segmented Landsat imagery and land cover data products, and
created a stacked ensemble to map the probability of shrubland at a high
resolution (30 m) across the study area. Models were evaluated across
multiple sensitivity/specificity thresholds to generate a range of map
outputs that can support future ground-truthing efforts as part of
research and stewardship activities. We provide new maps of marginal
cover types, such as invasive shrublands and degraded young forests, and
demonstrate how to leverage an information-rich but geographically
incomplete data source (LIDAR) for large-scale contiguous land cover
classification and mapping based on widely available and standardized
time-series imagery. Our results suggest that using LiDAR, where it
exists, to label data for model training may help distinguish shrublands
from other, optically similar land cover classes. We additionally
demonstrate the utility of using targeted, regional land cover products
to supplement national products for classes of regional interest.

\hypertarget{methods}{%
\section{Methods}\label{methods}}

\begin{figure}

{\centering \includegraphics{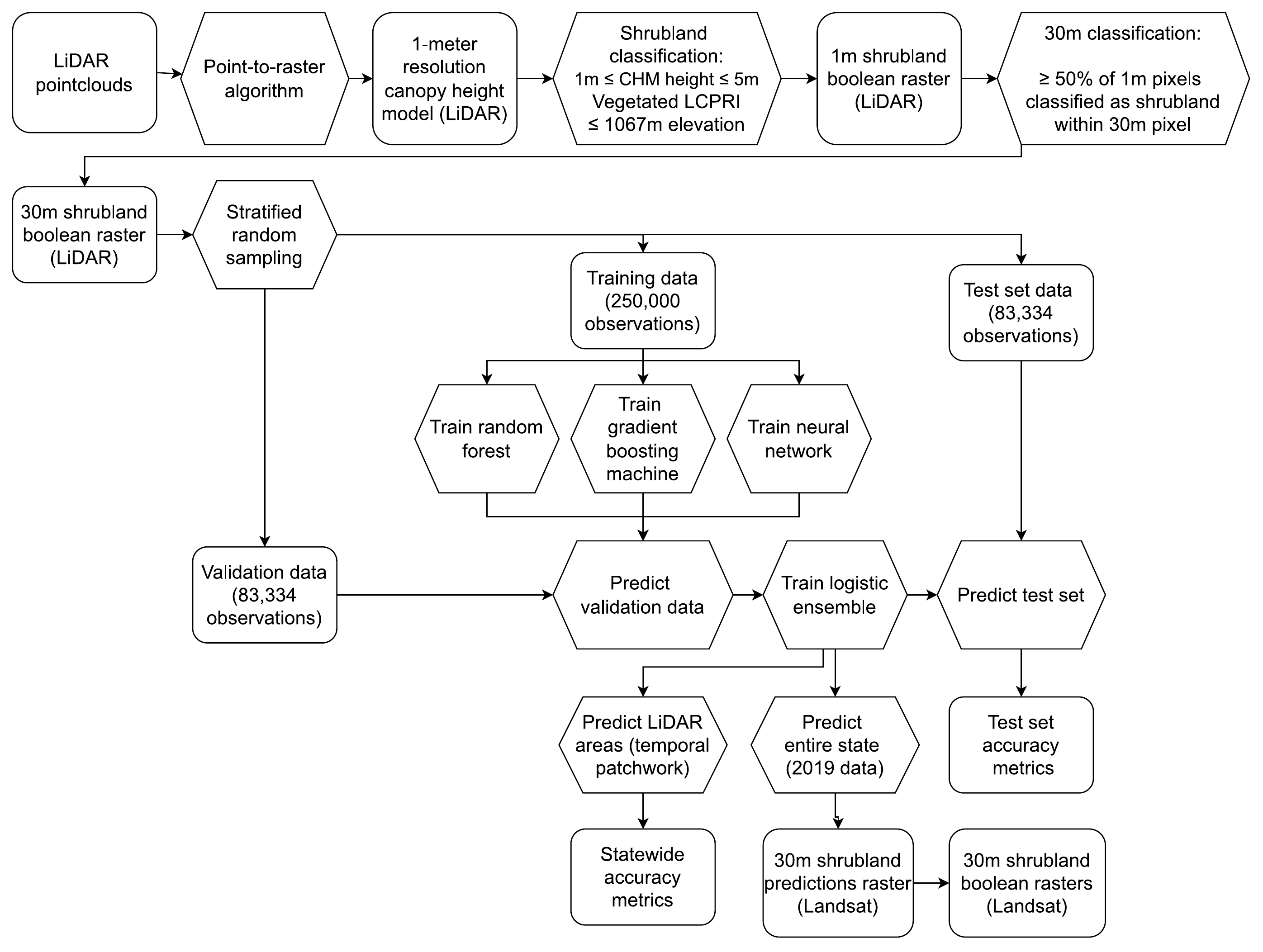}

}

\caption{\label{fig-flowchart}A flowchart diagram showing the key
elements of the identification methodology. Rectangular boxes represent
data products and results, while hexagonal boxes represent
methodological steps.}

\end{figure}

In order to map potential shrubland across New York State, we first
identified low-stature vegetation across a discontinuous temporal
patchwork of 1m resolution canopy height models (CHMs) derived from 19
distinct LiDAR acquisitions. We then aggregated these 1m identifications
into a 30m resolution raster surface, associated this surface with
multiple data products derived from temporally matching remote sensing
observations (including Landsat imagery as well as climate and
topographic data), and used this data to produce a stacked ensemble
model composed of a random forest, gradient boosting machine, and
artificial neural network combined through a logistic regression. We
used this ensemble model to predict shrubland for a spatiotemporal
patchwork matching LiDAR acquisitions, as well as for the entire state
in 2019. A flowchart of this process is included as
Figure~\ref{fig-flowchart}.

\hypertarget{study-area}{%
\subsection{Study area}\label{study-area}}

New York State spans an area of 141,297 \(\operatorname{km}^2\) of the
northeastern United States. Extensive land clearing for agriculture and
industry in the 18th through 19th centuries decimated forests throughout
the region, with forest cover dropping to 10-30\% of the landscape by
1880 (Lorimer 2001). While total forest cover recovered rapidly at the
turn of the 20th century, these forests were almost entirely young due
to the combination of regeneration on abandoned farmland with the
continual coppicing and harvesting of more established woodlots for
fuelwood and other products (Whitney 1994). A variety of factors, among
them a decrease in the use of wood for residential heating and an
increase in forest conservation efforts, caused these forests to begin
to mature in the 1930s, with the effect that the majority of forest
stands across the Northeast are now over 100 years old. Two forest
preserves, the Adirondack Park in the northeast and the Catskill Park in
the southeast, have been protected in the New York State constitution as
``forever wild''; timber harvesting has been generally prohibited on
state-owned parcels in these regions since they were incorporated into
the Forest Preserve, a process beginning in 1885. As a result, there is
very little shrubland in these preserves.

Elevations across New York State range from -2 m to 1,584 m above sea
level (U.S. Geological Survey 2019), with daily temperatures in 2019
ranging from -17 °C to 28 °C and monthly precipitation for the same
period ranging from 5.0 cm to 16.8 cm (NOAA National Centers for
Environmental Information 2022). The majority of the state occupies the
northern hardwoods-hemlock forest region, though there are important
inclusions of beech-maple-basswood and Appalachian oak communities in
the western and southern reaches of the state, respectively (Dyer 2006).

\hypertarget{lidar-data-and-shrubland-identification}{%
\subsection{LiDAR Data and Shrubland
Identification}\label{lidar-data-and-shrubland-identification}}

Although a distinction exists between early-successional and young
forest habitats in eastern North America, for simplicity we have
followed King and Schlossberg (2014) in combining these categories into
a single shrubland classification due to their structural similarity.
This terminology aligns with Anderson (1976), who described shrubland in
the eastern United States as ``former croplands or pasture lands
(cleared from original forest land) which now have grown up in brush in
transition back to forest land to the extent that they are no longer
identifiable as cropland or pasture from remote sensor imagery.''

\begin{figure}

{\centering \includegraphics[width=1\textwidth,height=\textheight]{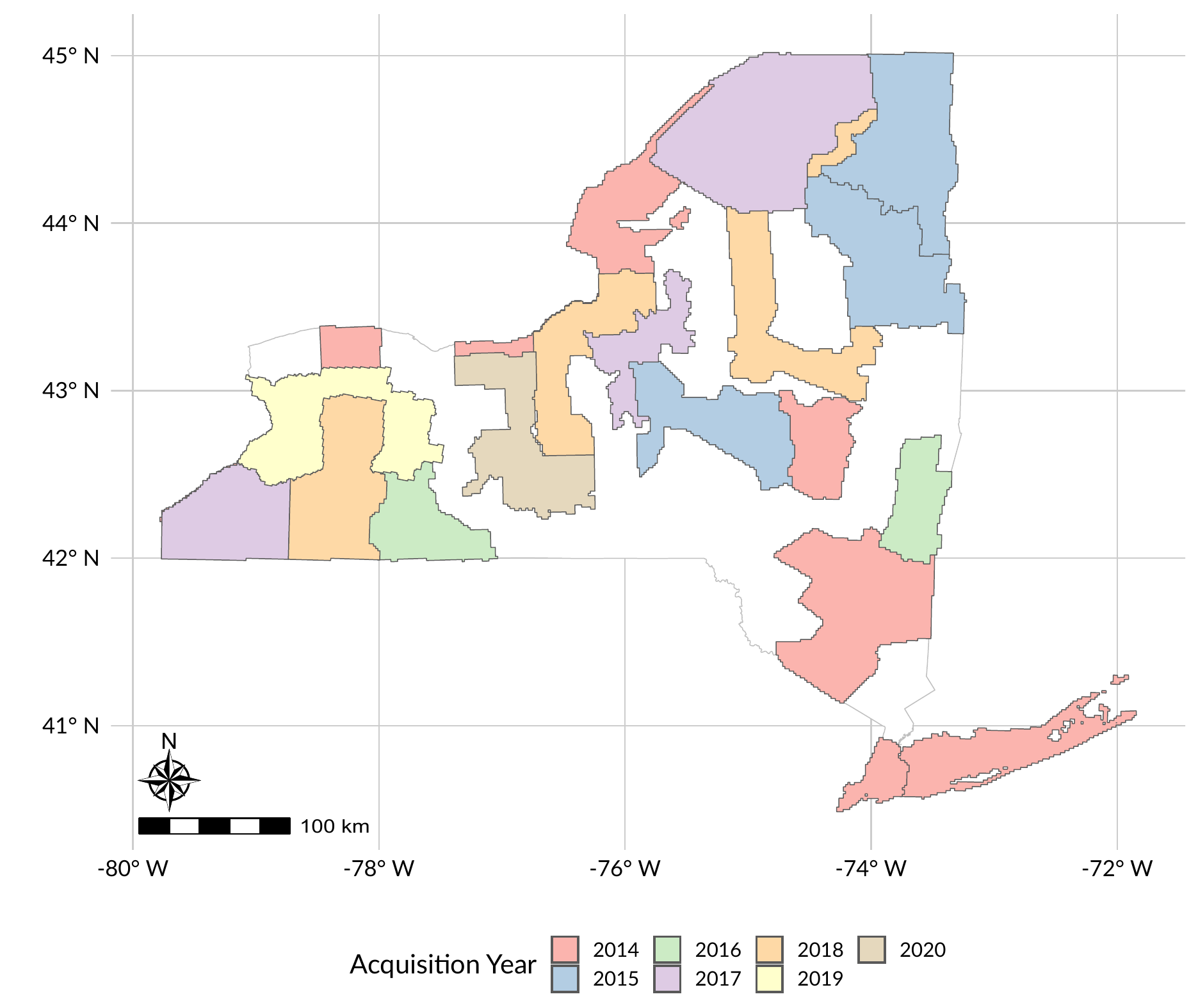}

}

\caption{\label{fig-lidarboundaries}Boundaries for all LiDAR coverages
used in this project, colored by year of data acquisition. More
information about each coverage is included as Online Resource 1.}

\end{figure}

Shrubland was identified using 19 distinct leaf-off LiDAR acquisitions,
collected and made freely available as part of the US Geological
Survey's 3D Elevation program. All LiDAR used in this study was
collected between 2014 and 2020, with point densities ranging from 1.98
to 3.24 points per square meter (Figure~\ref{fig-lidarboundaries}). All
LiDAR data had a vertical accuracy RMSE of \(\leq\) 10 cm. While
horizontal accuracy was not typically reported in provided LiDAR
metadata, horizontal RMSE for all data sets is expected to be \(\leq\)
68 cm (ASPRS 2014). More information about individual LiDAR coverages is
available as Supplementary Materials 1. These point clouds were
converted to 1 m canopy height models (CHMs) using a point-to-raster
algorithm implemented in the lidR R package (Roussel et al. 2020). To
reflect the 2D nature of a LiDAR return footprint, and mitigate
potential voids in the resulting CHM, each return was replaced with a
circle of returns with a diameter equal to the pulse width present in
the metadata (default 0.5 m; Supplementary Materials 1). These CHMs were
then masked to exclude any pixel assigned a non-vegetation primary land
cover classification by the temporally-matching USGS LCMAP land cover
product (namely the developed, water, ice and snow, and barren classes)
using the terra R package (Brown et al. 2020; Hijmans 2021; R Core Team
2021). CHMs were also masked to exclude any pixel with an elevation
above 1067m (3500 ft), as shorter canopy heights at these elevations
likely represent krumhholz (stunted trees near elevational treeline)
instead of shrubs or regenerating forest. Shrubland was then defined as
any 1 m pixel with a CHM height between 1-5 m. The lower threshold was
defined to avoid classifying cropland and human structures (such as low
walls) as shrubland, and the upper threshold defined to match the USGS
NLCD definition of shrubland, which was derived from the Anderson
classification system (Yang et al. 2018; Anderson et al. 1976). While
this coarse definition allows for efficient and automated labeling of
data, it also has its limitations, including being reliant upon the
accuracy and unbiasedness of LCMAP class predictions and potentially
incorrectly classifying some cropland, including orchards, as
shrublands. To evaluate these limitations, we validated model
predictions using manual inspection of optical imagery, described
further in Section~\ref{sec-mod-fit}.

These 1 m pixels were then aggregated into 30 m resolution pixels using
GDAL (GDAL/OGR contributors 2021), with each 30 m pixel defined as
shrubland if more than 50\% of its constituent 1 m subpixels were
identified as shrubland.

\hypertarget{predictor-creation}{%
\subsection{Predictor Creation}\label{predictor-creation}}

We produced a set of 10 annual Landsat-derived predictors by processing
Landsat analysis ready data (ARD, Dwyer et al. 2018) in Google Earth
Engine (GEE, Gorelick et al. 2017), using a medoid composite of imagery
acquired between July 1st and September 1st of the same year LiDAR was
acquired at each pixel. Landsat ARD was processed using the LandTrendr
implementation in GEE (hereafter LT-GEE) to fill gaps (e.g., clouds and
shadows) in the annual time series, smooth interannual variations
(noise), and quantify the disturbance history for each pixel (Kennedy,
Yang, and Cohen 2010; Kennedy et al. 2018). The LT-GEE predictors
included three tasseled cap indices, (brightness - TCB, greenness - TCG,
and wetness - TCW) and their respective deltas computed with a 1-year
lag, all fit to Normalized Burn Ratio (NBR) temporally segmented
vertices, as well as an NBR index and respective 1-year delta (Kauth and
Thomas 1976; Cocke, Fulé, and Crouse 2005; Kennedy et al. 2018). We also
processed a separate NBR segmented time-series with LT-GEE parameters
tailored to be more sensitive to the timing of discrete disturbances
using LT-GEE code to produce two predictors describing disturbances at
an individual pixel: year-of-most-recent-disturbance (YOD; 1985-2020)
and associated magnitude-of-most-recent-disturbance (MAG; unitless
measure of change in NBR value) (Kennedy et al. 2018). All predictors
are described in Table~\ref{tbl-predictors}. The 8 annual indices and
associated deltas aimed to capture the surface reflectance for a given
pixel at a given time, while the two disturbance predictors aimed to
describe disturbance history for a given pixel. NBR was chosen as the
base predictor, providing disturbance history information and temporal
break-points to which tasseled cap predictors were fit, as it has been
shown to be the most sensitive for capturing disturbance events
(Kennedy, Yang, and Cohen 2010). Information on LT-GEE parameters used
is included as Supplementary Materials 2.

\hypertarget{tbl-predictors}{}
\begin{table}[H]
\caption{\label{tbl-predictors}Definitions of predictors used for model fitting. }\tabularnewline

\centering
\begin{tabular}[t]{>{\raggedright\arraybackslash}p{10em}>{\raggedright\arraybackslash}p{22em}}
\toprule
\multicolumn{1}{c}{Predictor} & \multicolumn{1}{c}{Definition}\\
\midrule
TCB, TCW, TCG & Tassled cap brightness, wetness, and greenness, with noise removed using LT-GEE\\
\addlinespace
NBR & Normalized burn ratio with noise removed using LT-GEE\\
\addlinespace
MAG, YOD & Magnitude and year of most recent disturbance, as identified using LT-GEE\\
\addlinespace
PRECIP, TMAX, TMIN & 30-year normals for precipitation, maximum temperaure, and minimum temperature, derived from annual PRISM climate models\\
\addlinespace
ASPECT, ELEVATION, SLOPE, TWI & Aspect, elevation, slope, and topographic wetness index derived from a 30-meter digital elevation model\\
\addlinespace
LCSEC & LCMAP secondary land cover classification\\
\bottomrule
\end{tabular}
\end{table}

In addition to Landsat-derived predictors, a set of steady-state
ancillary predictors was included to represent geospatial variation in
climate and topography (Kennedy et al. 2018). These predictors included
precipitation and temperature 30 year normals derived from PRISM Climate
Group data (PRISM Climate Group 2022), the secondary land cover
classification prediction from LCMAP (Brown et al. 2020), and elevation,
aspect, slope, and topographic wetness indices derived from a 30 meter
digital elevation model (Mahoney, Beier, and Ackerman 2022; U.S.
Geological Survey 2019; Beven and Kirkby 1979). In total, models were
fit using 14 separate predictors (Table~\ref{tbl-predictors}).

\hypertarget{sec-mod-fit}{%
\subsection{Model Fitting and Evaluation}\label{sec-mod-fit}}

Each LiDAR-derived shrubland layer was combined with a set of temporally
matching predictors, which were then merged into a single temporal
patchwork data set representing each region of the state during its year
of LiDAR acquisition. A total of 416,668 pixels were then sampled from
this patchwork set (Figure~\ref{fig-flowchart}), stratified so that half
(208,334) represented shrubland and half other land cover types. These
pixels were then split at random into a training set of 250,000 pixels,
a validation set of 83,334 pixels, and a hold-out evaluation set of
83,334 pixels. We then fit three separate models, a random forest
(Breiman 2001), stochastic gradient boosting machine (Friedman 2002),
and deep neural network (LeCun, Bengio, and Hinton 2015), against the
training data set to estimate the probability of a given pixel
representing shrubland. Models were fit using hyperparameters chosen to
minimize out-of-sample binary cross entropy (Good 1952).

The random forest was fit using 3,000 trees, each using a random
bootstrap sample of 20\% of the data, sampled with replacement, a
minimum node size of 6 observations, a single random variable per split,
and splitting to minimize Gini impurity. The gradient boosting machine
was fit using 2,500 trees, each with unlimited depth, a maximum of 14
leaves, a learning rate of 0.01, a minimum of 10 observations per leaf
and 3 observations per bin, an L1 regularization constant of 0 and a L2
regularization constant of 0.5. Each tree was fit using a new bootstrap
sample with half the number of observations as the training data, with
access to 90\% of predictors. The neural net was a simple additive
neural network with seven layers: five densely connected layers with
numbers of nodes halving with each additional layer, decreasing from 256
to 128 to 64 to 32 to 16, then feeding into a 20\% dropout layer before
the final densely connected output layer with a single node. All dense
layers used rectified linear activation functions save for the output
layer, which used a sigmoid activation function. The model was trained
using 1,000 epochs, but final weights used the epoch which maximized the
area under the precision-recall curve of the validation data set.

We then used each of these models to predict the probability of a pixel
representing shrubland for each observation in the validation data set.
Next, we fit a logistic regression to the validation set predictors and
predicted probabilities to combine our three models into a single
stacked ensemble model (Wolpert 1992; Dormann et al. 2018). This
ensemble model was then used to generate predictions for the test set,
for the temporal patchwork data set, and for data reflecting the entire
state for 2019 (chosen in order to compare predictions to the 2019 NLCD
land cover map). The same model was used for predicting each of these
data sets.

Probability thresholds used to classify individual pixels were chosen
using model predictions for the validation set. Four separate thresholds
were identified: the one that maximized the model's summed sensitivity
(Equation~\ref{eq-sens}) and specificity (Equation~\ref{eq-spec}),
calculated using Youden's J statistic (hereafter `Youden Optimal')
(Youden 1950); and three that maximized sensitivity while keeping
specificity above 90\%, 95\%, and 99\%. All accuracy metrics were
assessed using the temporal patchwork data set, with Landsat-derived
predictors temporally matched to the LiDAR data set. All metrics were
calculated considering shrubland pixels as positive cases; higher
specificity targets reflected the relatively rare abundance of shrubland
throughout the state (approximately 2.5\% of mapped pixels)
necessitating low levels of false positives. Predictions against both
the test set and the temporal patchwork data set were classified using
each of these thresholds, then assessed using sensitivity
(Equation~\ref{eq-sens}), specificity (Equation~\ref{eq-spec}),
precision (Equation~\ref{eq-prec}), and F1 score (Equation~\ref{eq-f1}).

\begin{equation}\protect\hypertarget{eq-sens}{}{
\operatorname{Sensitivity} = \frac{T_{\operatorname{Positives}}}{T_{\operatorname{Positives}}+F_{\operatorname{Negatives}}}
}\label{eq-sens}\end{equation}

\begin{equation}\protect\hypertarget{eq-spec}{}{
\operatorname{Specificity} = \frac{T_{\operatorname{Negatives}}}{F_{\operatorname{Positives}}+T_{\operatorname{Negatives}}}
}\label{eq-spec}\end{equation}

\begin{equation}\protect\hypertarget{eq-prec}{}{
\operatorname{Precision} = \frac{T_{\operatorname{Positives}}}{T_{\operatorname{Positives}}+F_{\operatorname{Positives}}}
}\label{eq-prec}\end{equation}

\begin{equation}\protect\hypertarget{eq-f1}{}{
\operatorname{F1} = \frac{2*\operatorname{Precision}*\operatorname{Sensitivity}}{\operatorname{Precision}+\operatorname{Sensitivity}}
}\label{eq-f1}\end{equation}

Where \(T_{\operatorname{Positives}}\) is the number of pixels correctly
classified as shrubland, \(F_{\operatorname{Positives}}\) the number of
pixels incorrectly classified as shrubland,
\(T_{\operatorname{Negatives}}\) the number of pixels correctly
classified as not being shrubland, and \(F_{\operatorname{Negatives}}\)
the number of pixels incorrectly classified as not being shrubland.

Models were additionally assessed using the area under the receiver
operating characteristic curve (AUC) (Austin and Steyerberg 2012),
calculated for the test set using all observations and for the temporal
patchwork set using a random sample of 1,000,000 pixels due to
computational limitations. Given the imbalance of our classes, we do not
report overall accuracy or balanced accuracy for either the test set or
the temporal patchwork, given that approximately 97.5\% overall accuracy
could be achieved by never predicting shrubland.

A final assessment involved comparing a stratified sample of model
predictions to 1m resolution 2019 imagery from the National Agricultural
Inventory Program (US Department of Agriculture 2019). Predictions from
the 2019 ensemble model were stratified spatially into 51 16,974
\(\operatorname{km}^{2}\) hexagons (equaling an apothem of 70
kilometers). Within each hexagon, five pixels were randomly selected
from each of 12 probability bins (predicted probabilities from the
ensemble model between 0\% and 5\%, 5\% and 10\%, 90\% and 95\%, and
95\% and 100\%, as well as each decile between 10\% and 90\%). Samples
taken for two additional bins, ranging from 0\% to 1\% and 99\% to
100\%, were merged into the 0\% to 5\% and 95\% to 100\% bins due to the
rarity of these extreme probabilities, resulting in these bins having
slightly more observations. When regions of a hexagon extended beyond
the mapped area, the number of pixels selected per bin was scaled in
proportion to the mapped region of the hexagon. Pixels were classified
as either clearly representing shrubland, clearly representing a
non-shrubland class, or as ``unknown'' if the correct classification
could not be ascertained from imagery. A single rater classified all
validation pixels.

All models were fit using either R version 4.1.2 (R Core Team 2021) or
Python version 3.9.10 (Python Core Team 2022). Random forests were fit
using the ranger R package (Wright and Ziegler 2017), gradient boosting
machines using lightgbm (Ke et al. 2017), and neural nets using keras
(Chollet 2015).

\hypertarget{results}{%
\section{Results}\label{results}}

\hypertarget{lidar-classification}{%
\subsection{LiDAR Classification}\label{lidar-classification}}

\begin{figure}

{\centering \includegraphics[width=1\textwidth,height=\textheight]{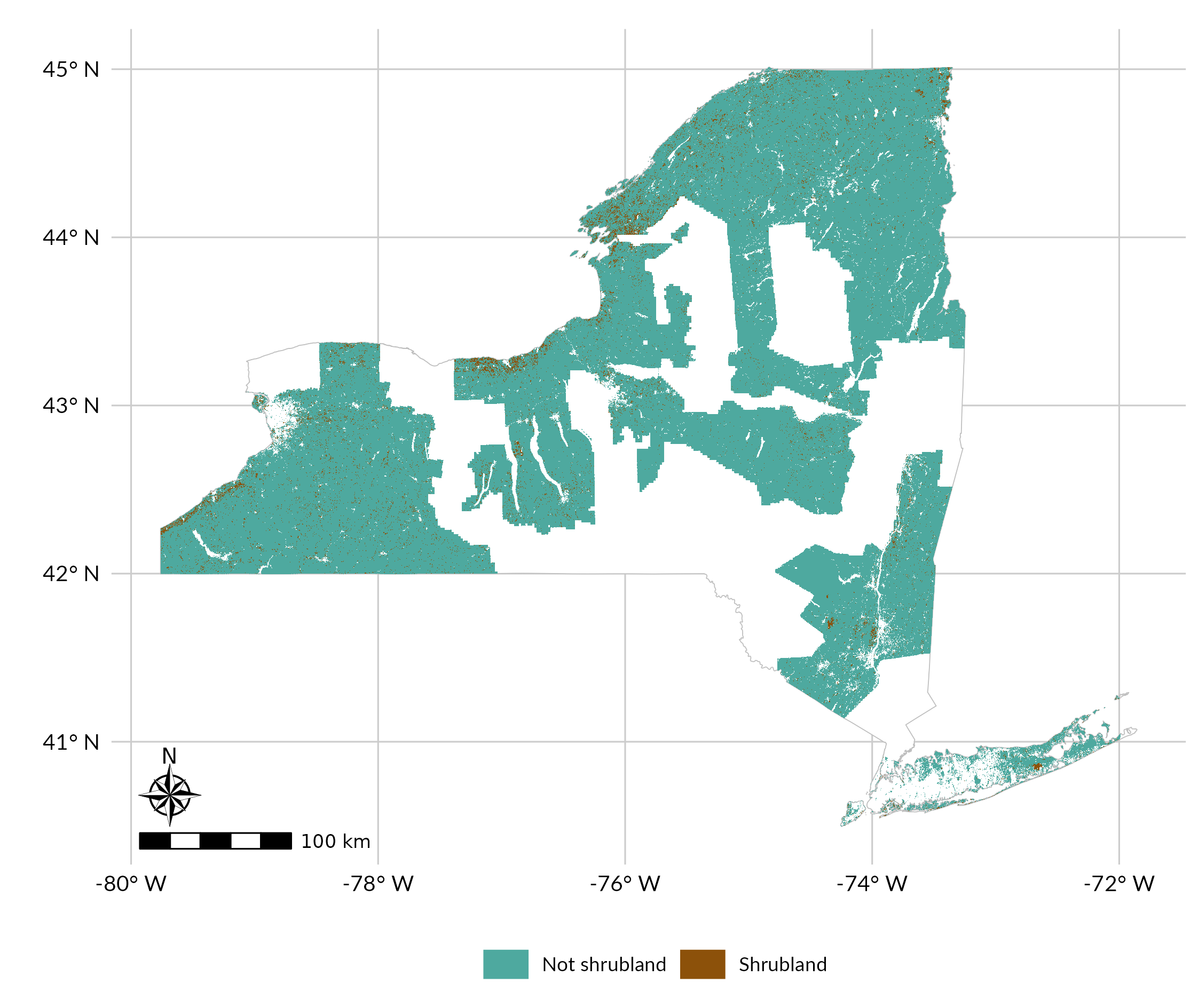}

}

\caption{\label{fig-lidarid}Identified shrubland areas within each
available LiDAR coverage. Shrubland was defined at a 1 meter resolution
as being any area within a vegetated LCPRI land cover class and below
1067 meters elevation with a LiDAR-derived height between 1 and 5
meters. 30 meter pixels, used for analysis and modeling, were then
defined as shrubland if more than 50\% of their contained 1 meter pixels
were classified as shrubland. In total, approximately 2.5\% of 30 meter
pixels were classified as shrubland.}

\end{figure}

Based on LIDAR-derived CHMs, approximately 2.5\% of the study area was
initially mapped as shrubland (1-5 m tall), representing about
\(1.83 x 10^6\) ha of the \(73.3 x 10^6\) ha land area that remained
after LCMAP masking removed non-vegetated cover types
(Figure~\ref{fig-lidarid}). Shrubland was identified in every LiDAR
coverage, with proportions ranging from 0.3\% (Great Gully) to 7.6\%
(Great Lakes) of the coverage footprint area (Figure~\ref{fig-lidarid}).

Shrubland cover was present in each vegetated LCMAP primary
classification classes, but was weighted more heavily towards areas
classified as cropland or tree cover (Table~\ref{tbl-lidarclass}).
Approximately 7.3\% of land classified by LCMAP as ``grassland/shrub''
was classified as shrubland through this process, though this result
should not be over-interpreted given the LCMAP ``grassland/shrub''
category includes herbaceous land covers alongside shrublands (Brown et
al. 2020).

\hypertarget{tbl-lidarclass}{}
\begin{table}
\caption{\label{tbl-lidarclass}Total area and amount of shrubland within each LCMAP primary land cover
classification (``LCPRI''), in square kilometers, for each map surface
(bolded text). Both the temporal patchwork and statewide models were
classified using a 95\% specificity threshold. }\tabularnewline

\centering
\begin{tabular}[t]{rrrr}
\toprule
\multicolumn{1}{c}{LCPRI} & \multicolumn{1}{c}{Total area} & \multicolumn{1}{c}{Shrubland area} & \multicolumn{1}{c}{\% Shrubland}\\
\midrule
\addlinespace[0.3em]
\multicolumn{4}{l}{\textbf{LiDAR Classification}}\\
\hspace{1em}Cropland & 20 077.6 & 679.5 & 3.4\%\\
\hspace{1em}Grass/Shrub & 2 085.3 & 151.8 & 7.3\%\\
\hspace{1em}Tree Cover & 44 879.5 & 597.2 & 1.3\%\\
\hspace{1em}Wetland & 6 635.0 & 467.9 & 7.1\%\\
\addlinespace[0.3em]
\multicolumn{4}{l}{\textbf{Temporal Patchwork}}\\
\hspace{1em}Cropland & 20 077.6 & 1 451.2 & 7.2\%\\
\hspace{1em}Grass/Shrub & 2 085.3 & 416.2 & 20.0\%\\
\hspace{1em}Tree Cover & 44 879.5 & 1 428.3 & 3.2\%\\
\hspace{1em}Wetland & 6 635.0 & 1 166.3 & 17.6\%\\
\addlinespace[0.3em]
\multicolumn{4}{l}{\textbf{2019 Statewide}}\\
\hspace{1em}Cropland & 29 922.9 & 2 005.9 & 6.7\%\\
\hspace{1em}Grass/Shrub & 2 994.2 & 549.3 & 18.3\%\\
\hspace{1em}Tree Cover & 69 419.0 & 1 694.6 & 2.4\%\\
\hspace{1em}Wetland & 9 008.3 & 1 425.0 & 15.8\%\\
\bottomrule
\end{tabular}
\end{table}

\hypertarget{model-accuracy}{%
\subsection{Model Accuracy}\label{model-accuracy}}

\begin{figure}

{\centering \includegraphics[width=1\textwidth,height=\textheight]{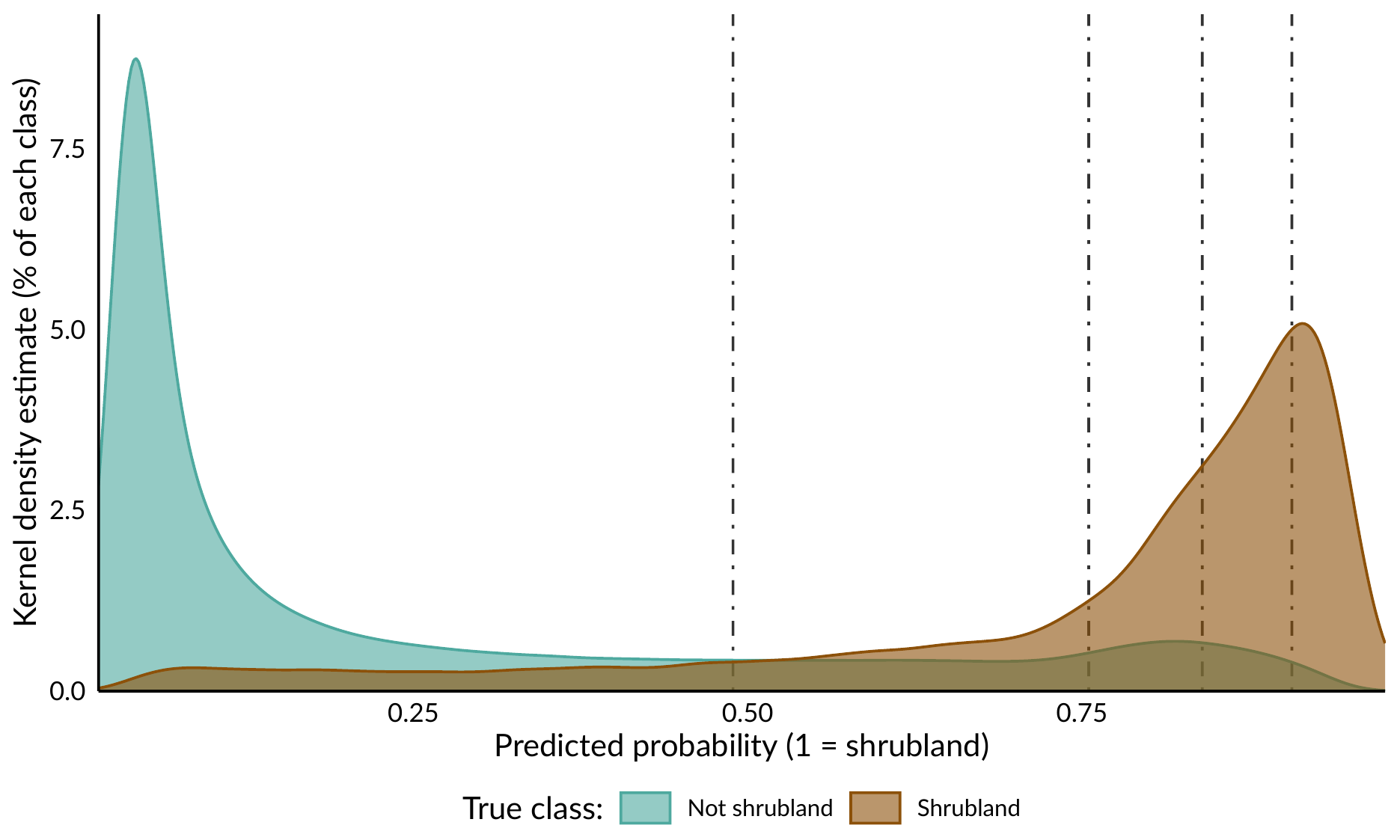}

}

\caption{\label{fig-predclass}Smoothed kernel density estimates of
predicted probability of shrubland for both shrubland and non-shrubland
pixels, calculated using a random sample of 1,000,000 pixels taken from
the LiDAR patchwork prediction surface using the logistic ensemble
model. Vertical lines indicate each of the four probability thresholds
used to classify pixels. Colors represent the correct classification of
the pixel.}

\end{figure}

The ensemble model had a higher AUC than the individual component models
on both the test set (Ensemble 0.893; random forest 0.843; gradient
boosting machine 0.889; neural network 0.883) and a random sample of
\(1 x 10^6\) pixels from the LiDAR temporal patchwork data set (Ensemble
0.904; random forest 0.844; gradient boosting machine 0.890; neural
network 0.901) (Figure~\ref{fig-predclass}; Supplementary Materials 3).
As a result, we focus here on accuracy assessments for the ensemble
model; accuracy assessments for the component models are included as
Supplementary Materials 3.

When evaluating models against the balanced test set, the highest F1
score (0.816) was achieved using the Youden-optimal classification
threshold (Table~\ref{tbl-modelacc}). The model retained its high AUC,
sensitivity, and specificity when predicting the LiDAR temporal
patchwork data set, precision was lower due to the large imbalance
between shrubland and other cover classes across the state. As a result,
the model attained its highest F1 score of 0.307 when using a
classification threshold to that targeted 95\% specificity.

\hypertarget{tbl-modelacc}{}
\begin{table}
\caption{\label{tbl-modelacc}Model accuracy metrics for logistic ensemble model with predictions
classified using various thresholds, calculated using both the balanced
test set and the LiDAR patchwork surface. AUC for the LiDAR patchwork
was calculated using a random sample of 1,000,000 pixels, while all
other metrics used all predicted pixels. Thresholds were selected using
a separate validation set, using values chosen to maximize the Youden J
statistic (`Youden optimal') or to target a certain minimum specificity
(`\% specificity'). }\tabularnewline

\centering
\begin{tabular}[t]{rrrrrr}
\toprule
\multicolumn{1}{c}{ } & \multicolumn{1}{c}{Threshold} & \multicolumn{1}{c}{Sensitivity} & \multicolumn{1}{c}{Specificity} & \multicolumn{1}{c}{Precision} & \multicolumn{1}{c}{F1}\\
\midrule
\addlinespace[0.3em]
\multicolumn{6}{l}{\textbf{Test set (AUC: 0.893)}}\\
\hspace{1em} &  &  &  &  \vphantom{1} & \\
\hspace{1em}Youden optimal & 0.489 & 0.842 & 0.780 & 0.791 & 0.816\\
\hspace{1em}90\% specificity & 0.755 & 0.659 & 0.900 & 0.867 & 0.807\\
\hspace{1em}95\% specificity & 0.840 & 0.496 & 0.949 & 0.906 & 0.641\\
\hspace{1em}99\% specificity & 0.907 & 0.218 & 0.989 & 0.952 & 0.355\\
\addlinespace[0.7em]
\multicolumn{6}{l}{\textbf{LiDAR patchwork (AUC: 0.904)}}\\
\hspace{1em} &  &  &  &  & \\
\hspace{1em}Youden optimal & 0.489 & 0.858 & 0.783 & 0.094 & 0.169\\
\hspace{1em}90\% specificity & 0.755 & 0.689 & 0.896 & 0.149 & 0.245\\
\hspace{1em}95\% specificity & 0.840 & 0.514 & 0.951 & 0.219 & 0.307\\
\hspace{1em}99\% specificity & 0.907 & 0.247 & 0.989 & 0.376 & 0.298\\
\bottomrule
\end{tabular}
\end{table}

\hypertarget{lidar-patchwork-predictions}{%
\subsubsection{LiDAR Patchwork
Predictions}\label{lidar-patchwork-predictions}}

\begin{figure}

{\centering \includegraphics[width=1\textwidth,height=\textheight]{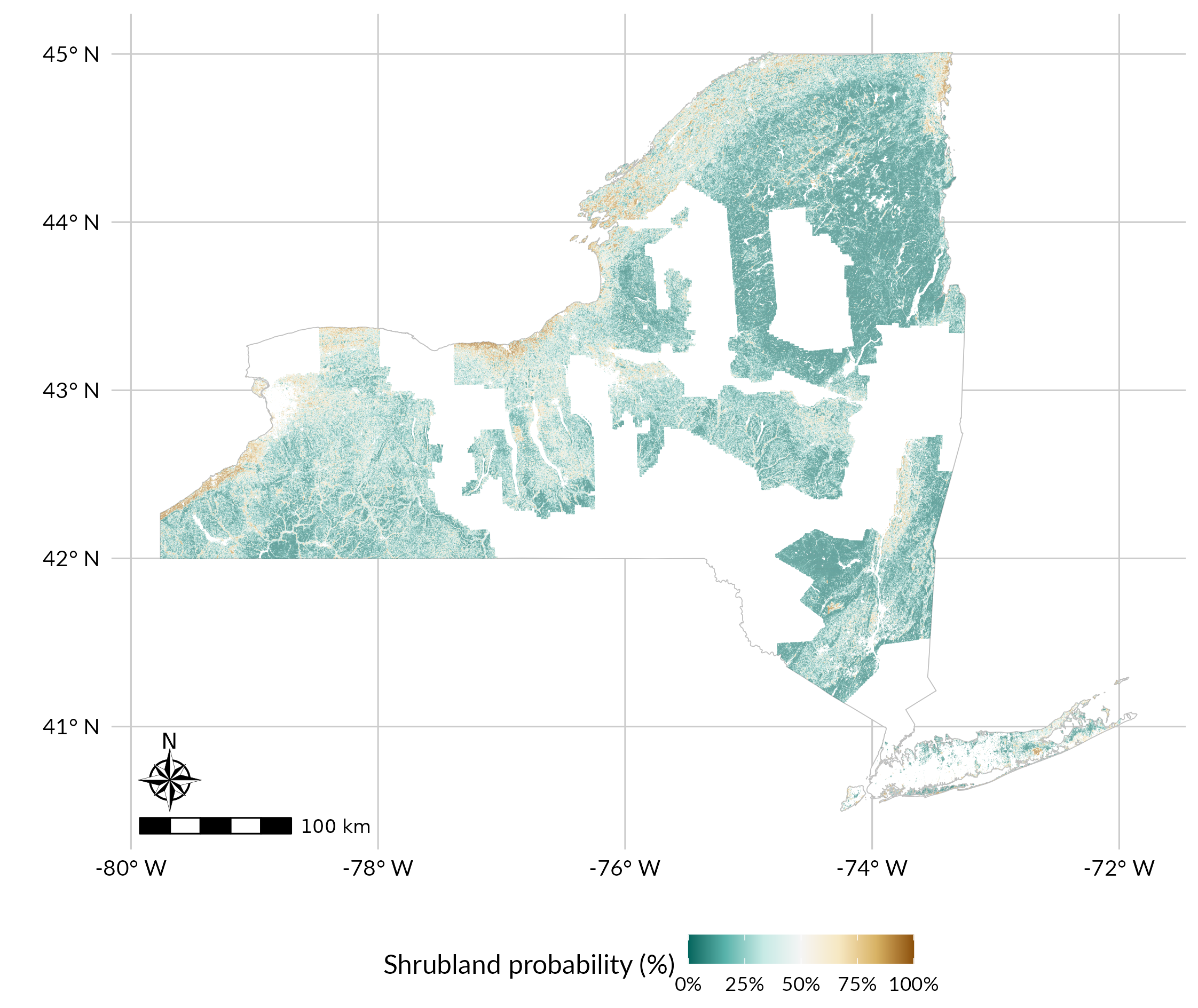}

}

\caption{\label{fig-lidarpred}Predicted probability of shrubland for the
boundaries of all used LiDAR coverages, from the logistic ensemble
model. Predictions were made using data reflecting the same year as
LiDAR acquisition; the map therefore represents a temporal patchwork of
predictions. Pixels in non-vegetated LCPRI land cover classes
(developed, water, ice/snow, and barren) or above 1067 meters in
elevation were not mapped and are shown in white.}

\end{figure}

When predicting the temporal patchwork, our model predicted the highest
probabilities of shrubland along the northern reaches of the state,
matching the distribution of shrubland in the true LiDAR-derived surface
(Figure~\ref{fig-lidarpred}). However, the model also predicted higher
than average probabilities of shrubland in the southwestern and central
regions of the state, neither of which were reflected in the original
LiDAR-derived surface.

\begin{figure}

{\centering \includegraphics[width=1\textwidth,height=\textheight]{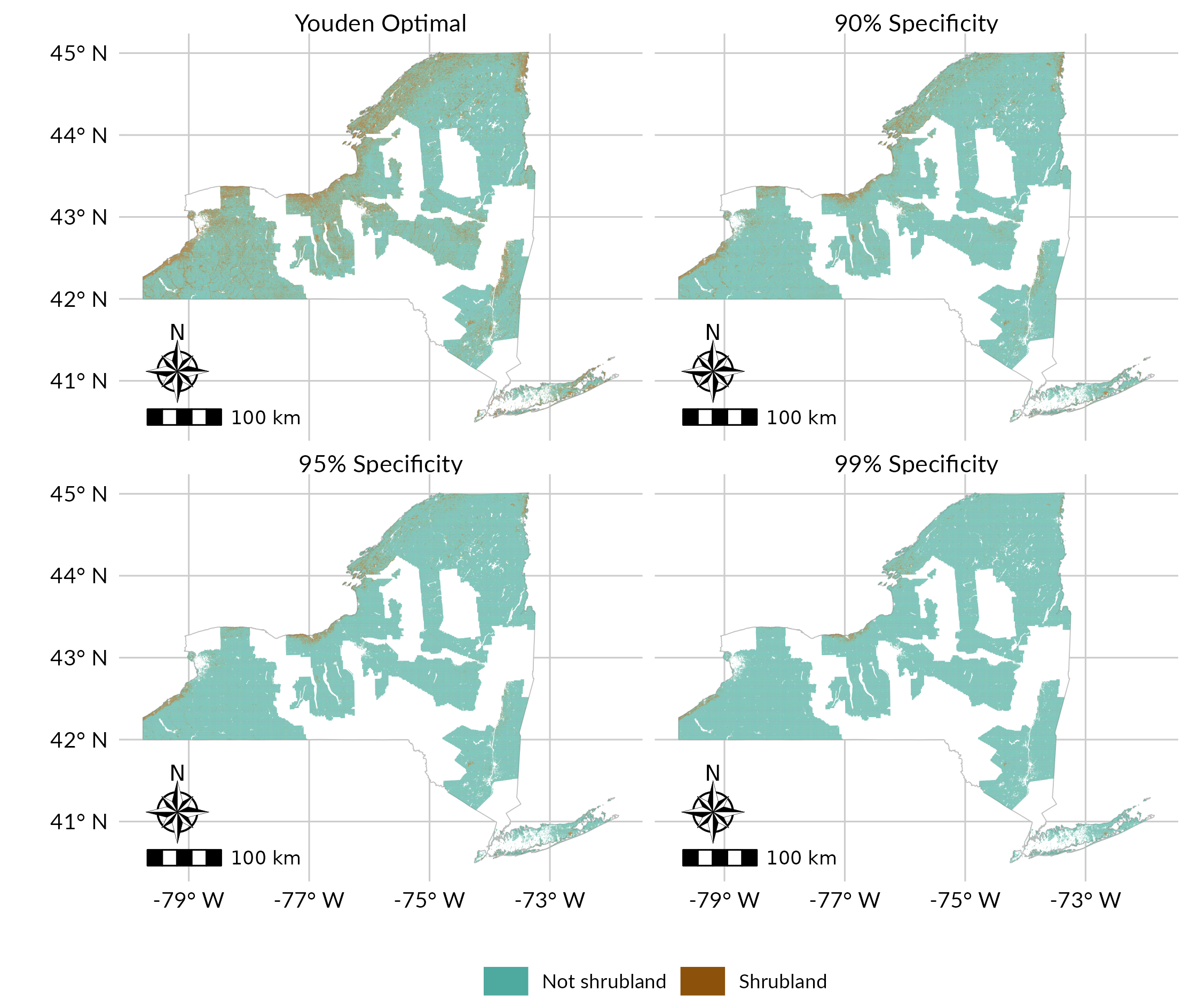}

}

\caption{\label{fig-lidarbool}Predicted shrubland locations within each
LiDAR coverage, from the logistic ensemble model. Predicted pixel
probabilities were classified using either the Youden-optimal threshold
(which maximizes both sensitivity and specificity) or a threshold chosen
to target a certain level of specificity, using thresholds derived from
the validation data set. Predictions were made using data reflecting the
same year as LiDAR acquisition; the map therefore represents a temporal
patchwork of predictions. Pixels in non-vegetated LCPRI land cover
classes (developed, water, ice/snow, and barren) or above 1067 meters in
elevation were not mapped and are shown in white.}

\end{figure}

Boolean surfaces classified using the Youden optimal probability
threshold classified 23\% of pixels as shrubland, an order of magnitude
greater than the 2.5\% identified from the true LiDAR-derived surface
(Figure~\ref{fig-lidarbool}). More conservative predictions based on
specificity thresholds predicted a lower proportion of shrubland (90\%
specificity: 11.9\%; 95\% specificity: 6.1\%, 99\% specificity: 1.7\%)
with a higher precision, resulting in more accurate overall predictions
and higher F1 scores.

\hypertarget{statewide-predictions}{%
\subsubsection{2019 Statewide Predictions}\label{statewide-predictions}}

\begin{figure}

{\centering \includegraphics[width=1\textwidth,height=\textheight]{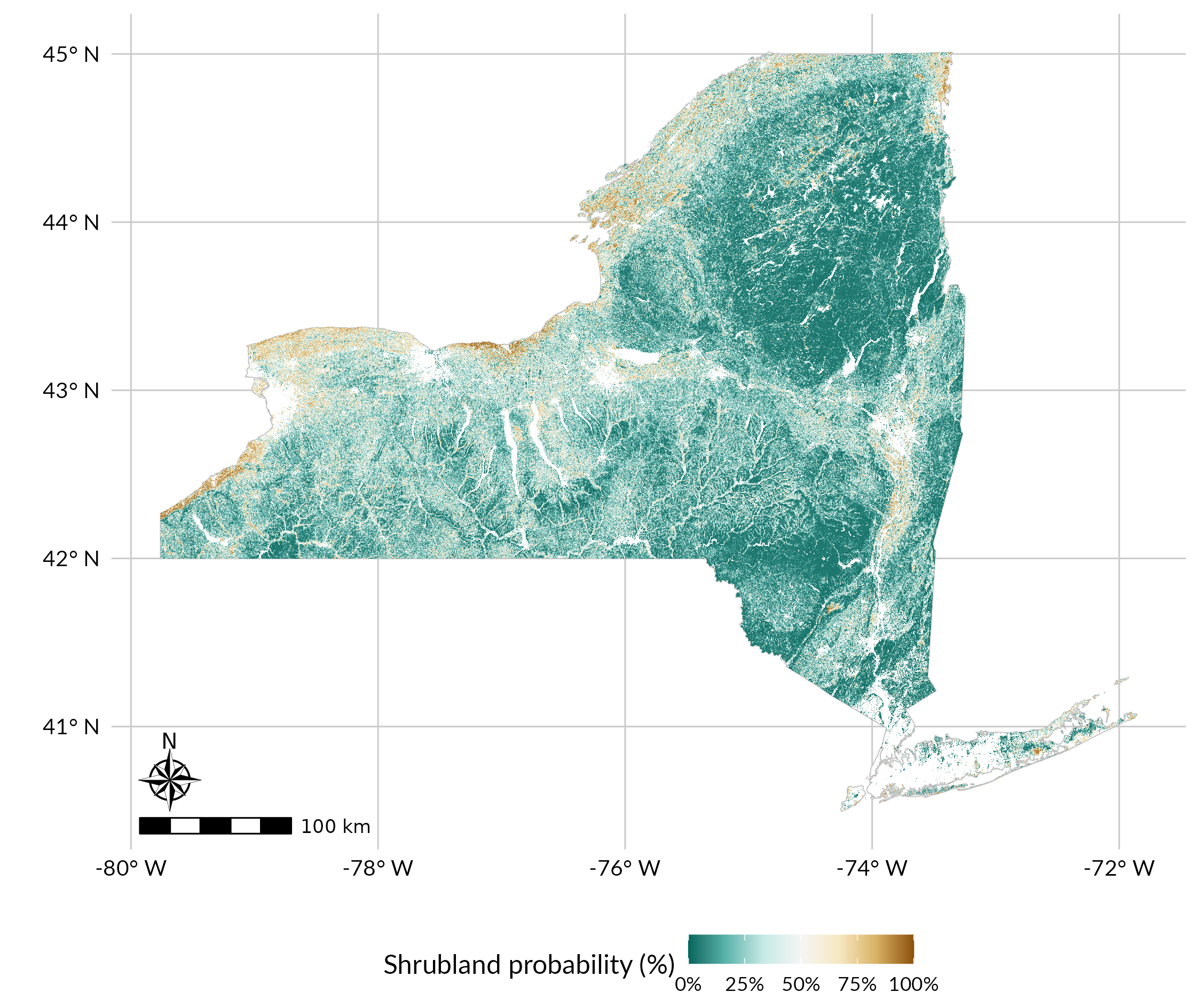}

}

\caption{\label{fig-landsatpred}Predicted probability of shrubland for
2019 across all mapped areas within New York State, from the logistic
ensemble model. Pixels in non-vegetated LCPRI land cover classes
(developed, water, ice/snow, and barren) or above 1067 meters in
elevation were not mapped and are shown in white.}

\end{figure}

Model predictions reflected a similar geographical distribution of
shrubland when extrapolating beyond the spatiotemporal boundaries of
available LiDAR data to map shrubland across the entire state for 2019.
Areas throughout the Adirondack Park and the Catskill Park, montane
regions with mostly contiguous forest cover, showed notably less
shrubland than more heavily populated areas
(Figure~\ref{fig-landsatpred}). As expected, predictions in areas
included in the LiDAR patchwork data set resembled the predictions for
the LiDAR patchwork surface (Figure~\ref{fig-lidarpred}), though with
some variation due in part to the temporal mismatch.

\begin{figure}

{\centering \includegraphics[width=1\textwidth,height=\textheight]{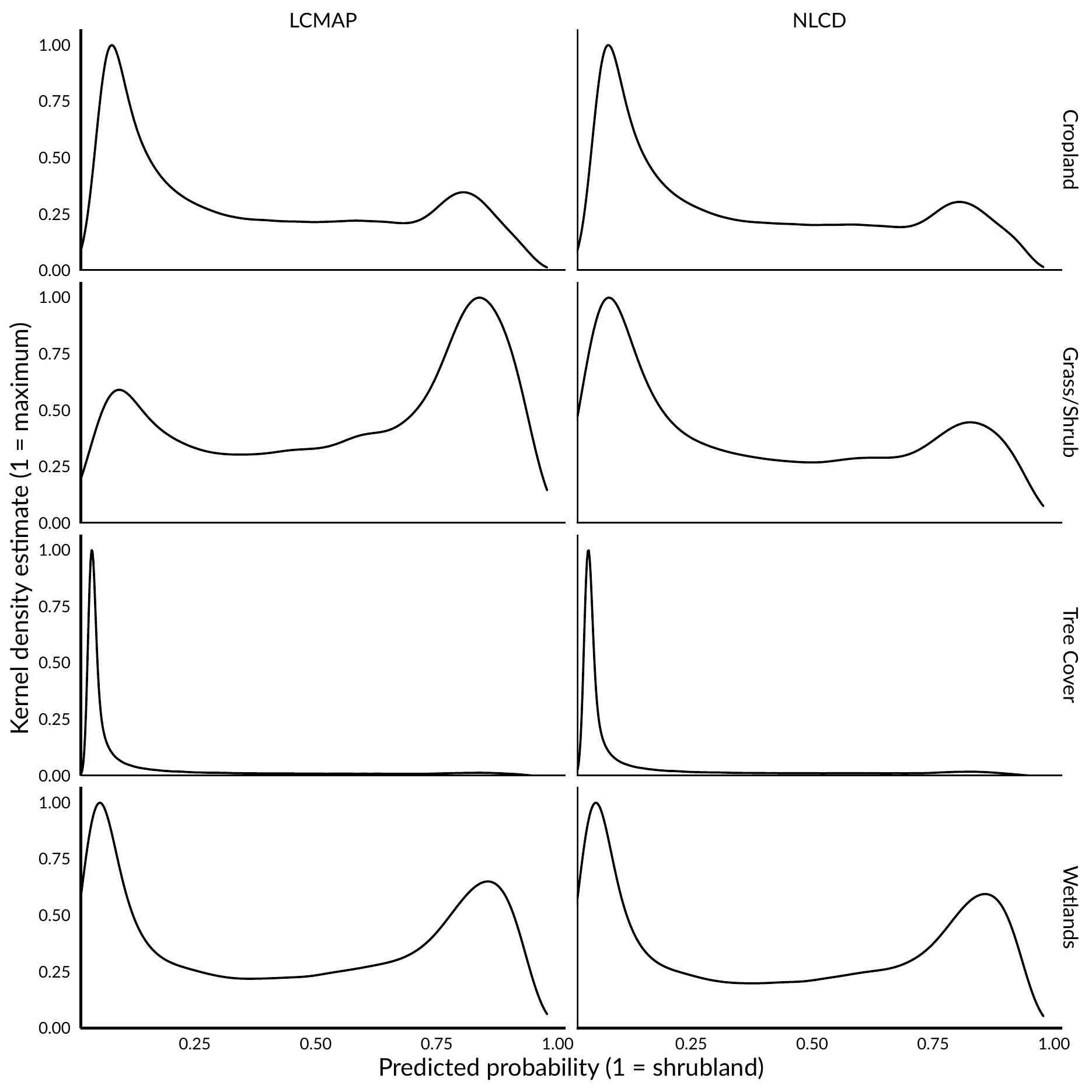}

}

\caption{\label{fig-lulcpred}Smoothed kernel density estimates of
predicted probability of shrubland for the included LCMAP classes,
calculated using a random sample of 1,000,000 pixels taken from the
LiDAR patchwork prediction surface using the logistic ensemble model.
Density estimates have been rescaled so that the most common probability
for each panel is assigned a value of 1. NLCD land cover classes were
remapped to LCMAP classes using LCMAP-defined translations.}

\end{figure}

Shrubland probabilities were highest in areas classified by 2019 LCMAP
as shrubland, as well as in areas classified as wetlands by either LCMAP
or NLCD (Figure~\ref{fig-lulcpred}). Areas classified as tree cover were
assigned extremely low probabilities.

\begin{figure}

{\centering \includegraphics[width=1\textwidth,height=\textheight]{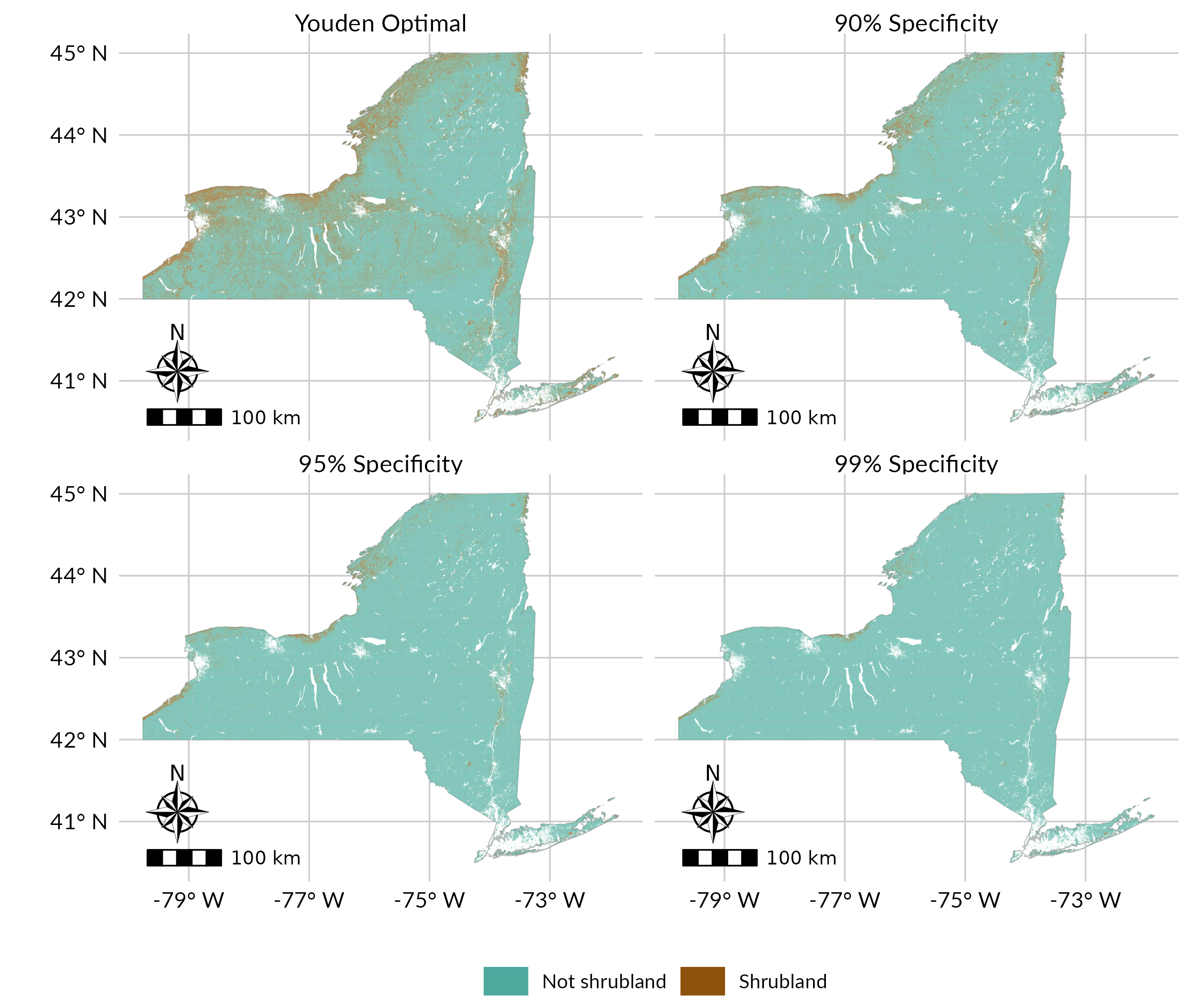}

}

\caption{\label{fig-landsatbool}Predicted shrubland locations for 2019
across the entire state, from the logistic ensemble model. Predicted
pixel probabilities were classified using either the Youden-optimal
threshold (which maximizes both sensitivity and specificity) or a
threshold chosen to target a certain level of specificity, using
thresholds derived from the validation data set. Pixels in non-vegetated
LCPRI land cover classes (developed, water, ice/snow, and barren) or
above 1067 meters in elevation were not mapped and are shown in white.}

\end{figure}

Predictions for 2019 classified using the Youden optimal probability
threshold classified 22.3\% of the state as shrubland
(Figure~\ref{fig-landsatbool}), in line with the Youden optimal
classified LiDAR patchwork data set. The target-specificity thresholds
classified more realistic proportions (90\% specificity: 10.7\%; 95\%
specificity: 5.1\%, 99\% specificity: 1.2\%).

\hypertarget{validation-using-2019-imagery}{%
\subsubsection{Validation Using 2019
Imagery}\label{validation-using-2019-imagery}}

Pixels with a higher predicted probability of shrubland were generally
more likely to represent shrubland, as determined using predictions and
imagery from 2019 (Table~\ref{tbl-validation}). Pixels with a nominal
predicted probability of shrubland at or above 0.95 were less likely to
actually represent shrubland than pixels with probabilities between 0.8
and 0.95, suggesting a failure to generalize to pixels with predictor
values outside the ranges reflected in the training data (Efron 2020)
(Table~\ref{tbl-validation}).

\hypertarget{tbl-validation}{}
\begin{table}
\caption{\label{tbl-validation}True classification of map pixels at various predicted shrubland
probabilities. Numbers in parentheses reflect the percentage of pixels
in each probability bracket in a given classification. Samples taken for
two additional bins, ranging from 0\% to 1\% and 99\% to 100\%, were
merged into the 0\% to 5\% and 95\% to 100\% bins due to the rarity of
these extreme probabilities, resulting in these bins having slightly
more observations. }\tabularnewline

\centering
\begin{tabular}[t]{rrrrr}
\toprule
\multicolumn{2}{c}{ } & \multicolumn{3}{c}{True classification (from imagery)} \\
\cmidrule(l{3pt}r{3pt}){3-5}
\multicolumn{1}{c}{Predicted probability of shrubland} & \multicolumn{1}{c}{\# in sample} & \multicolumn{1}{c}{Not shrubland} & \multicolumn{1}{c}{Unknown} & \multicolumn{1}{c}{Shrubland}\\
\midrule
(0,0.05] & 163 & 158 (96.9\%) & 5 (3.1\%) & 0 (0.0\%)\\
\addlinespace
(0.05,0.1] & 159 & 144 (90.6\%) & 11 (6.9\%) & 4 (2.5\%)\\
\addlinespace
(0.1,0.2] & 159 & 137 (86.2\%) & 18 (11.3\%) & 4 (2.5\%)\\
\addlinespace
(0.2,0.3] & 159 & 130 (81.8\%) & 17 (10.7\%) & 12 (7.5\%)\\
\addlinespace
(0.3,0.4] & 159 & 121 (76\%) & 19 (12\%) & 19 (12\%)\\
\addlinespace
(0.4,0.5] & 159 & 118 (74.2\%) & 24 (15.1\%) & 17 (10.7\%)\\
\addlinespace
(0.5,0.6] & 158 & 114 (72.2\%) & 26 (16.5\%) & 18 (11.4\%)\\
\addlinespace
(0.6,0.7] & 159 & 104 (65.4\%) & 25 (15.7\%) & 30 (18.9\%)\\
\addlinespace
(0.7,0.8] & 159 & 100 (62.9\%) & 24 (15.1\%) & 35 (22.0\%)\\
\addlinespace
(0.8,0.9] & 159 & 70 (44.0\%) & 25 (15.7\%) & 64 (40.3\%)\\
\addlinespace
(0.9,0.95] & 159 & 48 (30\%) & 30 (19\%) & 81 (51\%)\\
\addlinespace
(0.95,1] & 240 & 154 (64\%) & 16 (7\%) & 70 (29\%)\\
\bottomrule
\end{tabular}
\end{table}

\begin{figure}

{\centering \includegraphics{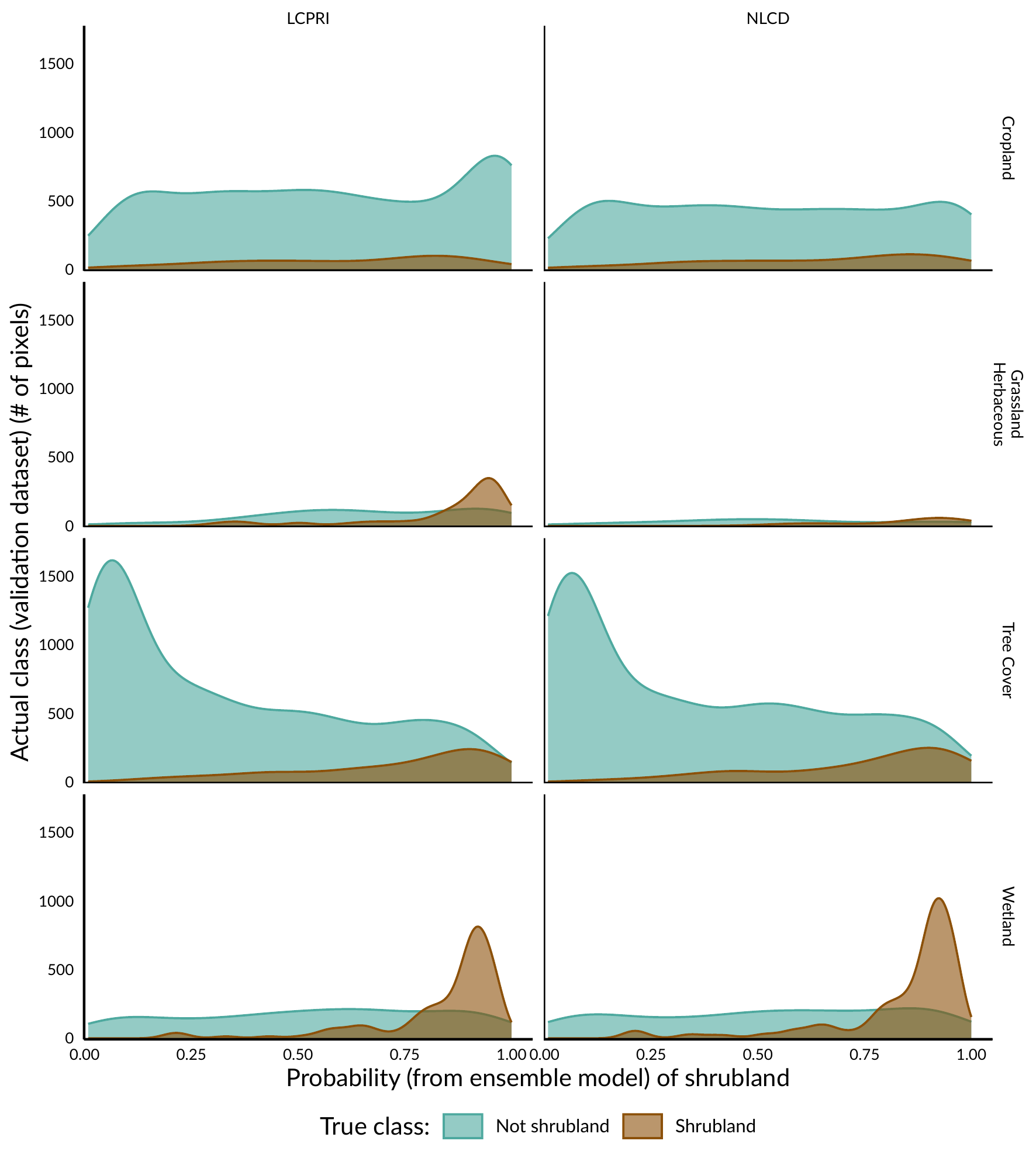}

}

\caption{\label{fig-pred-class-actual}Number of pixels (kernel density
estimate) representing shrubland (determined from NAIP imagery) by LCMAP
and NLCD classification. Pixels which could not be definitively
identified as shrubland are classed as ``not shrubland''.}

\end{figure}

The majority of shrubland pixels, as identified using imagery, were
classified as either tree cover or wetland pixels in both NLCD and LCMAP
(Figure~\ref{fig-pred-class-actual}). The majority of non-shrubland
pixels with a high predicted probability of shrubland were classified as
cropland by LCMAP (Figure~\ref{fig-pred-class-actual}).

\hypertarget{discussion}{%
\section{Discussion}\label{discussion}}

In this study, we produced a model capable of predicting shrubland
locations, as delineated using airborne LiDAR data and LCMAP land cover
classifications, across a large, fragmented and heterogenous landscape
(New York State). Overall, we found that our models were effective at
distinguishing between shrubland and other land cover classes, and
produced qualitatively sensible map outputs, even when extrapolating
beyond the original training data. Our results serve to demonstrate that
incorporating airborne LiDAR data can improve land cover
classifications, particularly for marginal, transitional and emergent
cover types that may not be well represented in land cover class
definitions. More practically, we provide new maps of marginal cover
types, such as invasive shrub/scrub and degraded young forests, that
have emerged mostly on post-agricultural and post-industrial lands
during the last century. While our model has limitations in the labeling
of training data and the inherent difficulty in predicting rare events,
it was effective at distinguishing between shrubland and other cover
types. This modeling approach addresses a persistent measurement gap and
enables monitoring, research, and stewardship of these emerging and
novel communities at a landscape scale.

\hypertarget{model-predictions-reflect-known-patterns}{%
\subsection{Model predictions reflect known
patterns}\label{model-predictions-reflect-known-patterns}}

Our model predictions, both for areas included in the LiDAR patchwork
and for the 2019 surface, reflect known patterns in land cover
throughout New York State. Although areas classified as developed by
LCMAP were excluded from predictions, areas of higher-intensity human
land use -- such as the Hudson Valley in the eastern region of the
state, the I-90 highway corridor running East-West (along roughly 43N
latitude), and the northern border of the state, particularly around the
Great Lakes -- were consistently classified as having a higher
probability of shrubland (Figure~\ref{fig-lidarpred},
Figure~\ref{fig-landsatpred}). These areas have likely been more
recently impacted by human activity, with cropland only more recently
being left to natural regeneration. By the same pattern, areas of lower
population density and less intensive land use history such as the
Adirondack and Catskill Parks have consistently low probabilities of
shrubland, reflecting the older, mid-successional forests that
characterize these areas. When extrapolating beyond the spatiotemporal
boundaries of available LiDAR data, our 2019 statewide model predicted a
similar abundance of shrubland with a similar regional distribution as
the LiDAR patchwork data set, with similar areas of shrubland along the
Great Lakes and adjacent to human population centers. This pattern
reflects the continuing decline in agricultural land across New York
State (USDA National Agricultural Statistics Service 2019); while much
agricultural land is being converted to developed land classes, a large
proportion of former cultivated lands have also been allowed to
regenerate.

\hypertarget{predicting-rare-events-is-challenging}{%
\subsection{Predicting rare events is
challenging}\label{predicting-rare-events-is-challenging}}

As previously noted, shrubland is rare in New York due to the state's
history of deforestation and regeneration combined with the relatively
mild disturbance regime of the northern forest (Lorimer 2001). One
challenge in predicting rare events is that even a low percentage of
false positives can quickly drown out true positives; a model with 90\%
specificity predicting a data set with only 1\% positive cases will
produce 9 false positives for every true positive it generates.

For this reason, we classified our predictions using a range of
thresholds targeting increasingly high specificities
(Table~\ref{tbl-modelacc}). These more stringent thresholds successfully
increased model precision (sometimes referred to as positive predictive
power), making it more likely that a positive prediction represents a
true positive, though at the cost of lower sensitivity. Of these three
thresholds, the 95\% specificity target best balanced sensitivity and
precision, based on its F1 score. Using this threshold, more than half
of true shrubland pixels were correctly classified and 22\% of positive
predictions reflect true shrubland (as defined in the LiDAR analysis),
approximately 10 times better than simply guessing using the shrubland
occurrence rate of 2.5\%. Although there is room for improvement, this
level of accuracy is sufficient to identify potential areas for further
research and stewardship.

Additionally, many of the areas identified as shrubland by our model and
confirmed via NAIP imagery were classified as tree cover or wetlands by
LCMAP and NLCD (Figure~\ref{fig-pred-class-actual}). This suggests that
incorporating LiDAR data in land cover mapping workflows may help to
distinguish shrublands from optically similar classes, even if LiDAR is
only used to improve the quality of training set labels. It also
potentially speaks to the benefits of more targeted, regional land cover
products to supplement the well-established national models; regional
efforts that can take advantage of regional data sets to improve
accuracy on cover types of regional importance.

\hypertarget{conclusion}{%
\section{Conclusion}\label{conclusion}}

This study aimed to predict the locations of shrubland across New York
State, in order to improve both understanding and stewardship of these
plant communities. Using a stacked ensemble model combining multiple
machine learning models fit to data labeled using a combination of
airborne LiDAR data and national land cover products, we generated
predictions of shrubland occurrences for both all spatiotemporal extents
with matching LiDAR data and for the entirety of New York State for
2019. Our model was highly effective at distinguishing between shrubland
and other cover types on both the test set (AUC 0.893) and the LiDAR
temporal patchwork (AUC 0.904), and balanced sensitivity and precision
effectively given the rarity of shrubland across the state. These
results suggest that combining remote sensing data from multiple sources
may improve land cover models, that regionally focused land cover models
may complement national products, and that shrubland may be effectively
identified and monitored using spaceborne remote sensing data.

\hypertarget{acknowledgments}{%
\section{Acknowledgments}\label{acknowledgments}}

Funding was provided by the Environmental Protection Fund via the NYS
Department of Environmental Conservation.

\hypertarget{disclosure-statement}{%
\section{Disclosure Statement}\label{disclosure-statement}}

The authors report there are no competing interests to declare.

\hypertarget{data-availability-statement}{%
\section{Data Availability
Statement}\label{data-availability-statement}}

Data is available online from https://doi.org/10.5281/zenodo.6519232.

\newpage{}

\hypertarget{references}{%
\section*{References}\label{references}}
\addcontentsline{toc}{section}{References}

\hypertarget{refs}{}
\begin{CSLReferences}{1}{0}
\leavevmode\vadjust pre{\hypertarget{ref-Alexander2015}{}}%
Alexander, Jake M, Jeffrey M Diaz, and Jonathan M Levine. 2015. {``Novel
Competitors Shape Species' Responses to Climate Change.''} \emph{Nature}
525: 515--18. \url{https://doi.org/10.1038/nature14952}.

\leavevmode\vadjust pre{\hypertarget{ref-Anderson1976}{}}%
Anderson, James R, Ernest E Hardy, John T Roach, and Richard E Witmer.
1976. \emph{A Land Use and Land Cover Classification System for Use with
Remote Sensor Data}. Vol. 964. US Government Printing Office.

\leavevmode\vadjust pre{\hypertarget{ref-Askins2001}{}}%
Askins, Robert A. 2001. {``Sustaining Biological Diversity in Early
Successional Communities: The Challenge of Managing Unpopular
Habitats.''} \emph{Wildlife Society Bulletin} 29 (2): 407--12.

\leavevmode\vadjust pre{\hypertarget{ref-ASPRS2014}{}}%
ASPRS. 2014. {``ASPRS Positional Accuracy Standards for Digital
Geospatial Data.''} \emph{Photogrammetric Engineering and Remote
Sensing} 81 (3): 53. \url{https://doi.org/10.14358/PERS.81.3.A1-A26}.

\leavevmode\vadjust pre{\hypertarget{ref-Austin2012}{}}%
Austin, Peter C, and Ewout W Steyerberg. 2012. {``Interpreting the
Concordance Statistic of a Logistic Regression Model: Relation to the
Variance and Odds Ratio of a Continuous Explanatory Variable.''}
\emph{BMC Medical Research Methodology} 12 (82).
\url{https://doi.org/10.1186/1471-2288-12-82}.

\leavevmode\vadjust pre{\hypertarget{ref-Benjamin2005}{}}%
Benjamin, Karyne, Gerald Domon, and Andre Bouchard. 2005. {``Vegetation
Composition and Succession of Abandoned Farmland: Effects of Ecological,
Historical and Spatial Factors.''} \emph{Landscape Ecology} 20 (6):
627--47. \url{https://doi.org/10.1007/s10980-005-0068-2}.

\leavevmode\vadjust pre{\hypertarget{ref-Beven1979}{}}%
Beven, Keith J., and Mike J. Kirkby. 1979. {``A Physically Based,
Variable Contributing Area Model of Basin Hydrology.''}
\emph{Hydrological Sciences Bulletin} 24 (1): 43--69.
\url{https://doi.org/10.1080/02626667909491834}.

\leavevmode\vadjust pre{\hypertarget{ref-bogner2018}{}}%
Bogner, Christina, Bumsuk Seo, Dorian Rohner, and Björn Reineking. 2018.
{``Classification of Rare Land Cover Types: Distinguishing Annual and
Perennial Crops in an Agricultural Catchment in South Korea.''} Edited
by Krishna Prasad Vadrevu. \emph{PLOS ONE} 13 (1): e0190476.
\url{https://doi.org/10.1371/journal.pone.0190476}.

\leavevmode\vadjust pre{\hypertarget{ref-Breiman2001}{}}%
Breiman, Leo. 2001. {``{Random Forests}.''} \emph{Machine Learning} 45:
5--32. \url{https://doi.org/10.1023/A:1010933404324}.

\leavevmode\vadjust pre{\hypertarget{ref-LCMAP}{}}%
Brown, Jesslyn F., Heather J. Tollerud, Christopher P. Barber, Qiang
Zhou, John L. Dwyer, James E. Vogelmann, Thomas R. Loveland, et al.
2020. {``Lessons Learned Implementing an Operational Continuous United
States National Land Change Monitoring Capability: The Land Change
Monitoring, Assessment, and Projection (LCMAP) Approach.''} \emph{Remote
Sensing of Environment} 238: 111356.
\url{https://doi.org/10.1016/j.rse.2019.111356}.

\leavevmode\vadjust pre{\hypertarget{ref-keras}{}}%
Chollet, François. 2015. {``Keras.''} \url{https://keras.io}.

\leavevmode\vadjust pre{\hypertarget{ref-Cocke2005}{}}%
Cocke, Allison E., Peter Z. Fulé, and Joseph E. Crouse. 2005.
{``Comparison of Burn Severity Assessments Using Differenced Normalized
Burn Ratio and Ground Data.''} \emph{International Journal of Wildland
Fire} 14: 189--98. \url{https://doi.org/10.1071/WF04010}.

\leavevmode\vadjust pre{\hypertarget{ref-Cramer2008}{}}%
Cramer, Viki A, Richard J Hobbs, and Rachel J Standish. 2008. {``What's
New about Old Fields? Land Abandonment and Ecosystem Assembly.''}
\emph{Trends in Ecology \& Evolution} 23 (2): 104--12.
\url{https://doi.org/10.1016/j.tree.2007.10.005}.

\leavevmode\vadjust pre{\hypertarget{ref-Dey2019}{}}%
Dey, Daniel C, Benjamin O Knapp, Mike A Battaglia, Robert L Deal, Justin
L Hart, Kevin L O'Hara, Callie J Schweitzer, and Thomas M Schuler. 2019.
{``Barriers to Natural Regeneration in Temperate Forests Across the
USA.''} \emph{New Forests} 50 (1): 11--40.
\url{https://doi.org/10.1007/s11056-018-09694-6}.

\leavevmode\vadjust pre{\hypertarget{ref-Dormann2018}{}}%
Dormann, Carsten F., Justin M. Calabrese, Gurutzeta Guillera-Arroita,
Eleni Matechou, Volker Bahn, Kamil Bartoń, Colin M. Beale, et al. 2018.
{``Model Averaging in Ecology: A Review of Bayesian,
Information-Theoretic, and Tactical Approaches for Predictive
Inference.''} \emph{Ecological Monographs} 88 (4): 485--504.
\url{https://doi.org/10.1002/ecm.1309}.

\leavevmode\vadjust pre{\hypertarget{ref-ARD2018}{}}%
Dwyer, John L., David P. Roy, Brian Sauer, Calli B. Jenkerson, Hankui K.
Zhang, and Leo Lymburner. 2018. {``Analysis Ready Data: Enabling
Analysis of the Landsat Archive.''} \emph{Remote Sensing} 10 (9).
\url{https://doi.org/10.3390/rs10091363}.

\leavevmode\vadjust pre{\hypertarget{ref-Dyer2006}{}}%
Dyer, James M. 2006. {``Revisiting the Deciduous Forests of Eastern
North America.''} \emph{BioScience} 56 (4): 341--52.
\url{https://doi.org/10.1641/0006-3568(2006)56\%5B341:RTDFOE\%5D2.0.CO;2}.

\leavevmode\vadjust pre{\hypertarget{ref-efron2020}{}}%
Efron, Bradley. 2020. {``Prediction, Estimation, and Attribution.''}
\emph{Journal of the American Statistical Association} 115 (530):
636--55. \url{https://doi.org/10.1080/01621459.2020.1762613}.

\leavevmode\vadjust pre{\hypertarget{ref-Falkowski2009}{}}%
Falkowski, Michael J., Jeffrey S. Evans, Sebastian Martinuzzi, Paul E.
Gessler, and Andrew T. Hudak. 2009. {``Characterizing Forest Succession
with Lidar Data: An Evaluation for the Inland Northwest, USA.''}
\emph{Remote Sensing of Environment} 113 (5): 946--56.
https://doi.org/\url{https://doi.org/10.1016/j.rse.2009.01.003}.

\leavevmode\vadjust pre{\hypertarget{ref-Fargione2018}{}}%
Fargione, Joseph E, Steven Bassett, Timothy Boucher, Scott D Bridgham,
Richard T Conant, Susan C Cook-Patton, Peter W Ellis, et al. 2018.
{``Natural Climate Solutions for the United States.''} \emph{Science
Advances} 4 (11): eaat1869.
\url{https://doi.org/10.1126/sciadv.aat1869}.

\leavevmode\vadjust pre{\hypertarget{ref-Flinn2005}{}}%
Flinn, Kathryn M, and Mark Vellend. 2005. {``Recovery of Forest Plant
Communities in Post-Agricultural Landscapes.''} \emph{Frontiers in
Ecology and the Environment} 3 (5): 243--50.
\url{https://doi.org/10.1890/1540-9295(2005)003\%5B0243:ROFPCI\%5D2.0.CO;2}.

\leavevmode\vadjust pre{\hypertarget{ref-Flinn2005a}{}}%
Flinn, Kathryn M, Mark Vellend, and PL Marks. 2005. {``Environmental
Causes and Consequences of Forest Clearance and Agricultural Abandonment
in Central New York, USA.''} \emph{Journal of Biogeography} 32 (3):
439--52. \url{https://doi.org/10.1111/j.1365-2699.2004.01198.x}.

\leavevmode\vadjust pre{\hypertarget{ref-Foster1998}{}}%
Foster, David R, Glenn Motzkin, and Benjamin Slater. 1998. {``Land-Use
History as Long-Term Broad-Scale Disturbance: Regional Forest Dynamics
in Central New England.''} \emph{Ecosystems} 1 (1): 96--119.
\url{https://doi.org/10.1007/s100219900008}.

\leavevmode\vadjust pre{\hypertarget{ref-Fridley2012}{}}%
Fridley, Jason. 2012. {``Extended Leaf Phenology and the Autumn Niche in
Deciduous Forest Invasions.''} \emph{Nature} 485: 359--62.
\url{https://doi.org/10.1038/nature11056}.

\leavevmode\vadjust pre{\hypertarget{ref-Friedman2002}{}}%
Friedman, Jerome H. 2002. {``Stochastic Gradient Boosting.''}
\emph{Computational Statistics and Data Analysis} 38 (4): 367--78.
\url{https://doi.org/10.1016/S0167-9473(01)00065-2}.

\leavevmode\vadjust pre{\hypertarget{ref-GDAL}{}}%
GDAL/OGR contributors. 2021. \emph{{GDAL/OGR} Geospatial Data
Abstraction Software Library}. Open Source Geospatial Foundation.
\url{https://gdal.org}.

\leavevmode\vadjust pre{\hypertarget{ref-Good1952}{}}%
Good, Irving J. 1952. {``Rational Decisions.''} \emph{Journal of the
Royal Statistical Society. Series B (Methodological)} 14 (1): 107--14.
\url{http://www.jstor.org/stable/2984087}.

\leavevmode\vadjust pre{\hypertarget{ref-Gorelick2017}{}}%
Gorelick, Noel, Matt Hancher, Mike Dixon, Simon Ilyushchenko, David
Thau, and Rebecca Moore. 2017. {``Google Earth Engine: Planetary-Scale
Geospatial Analysis for Everyone.''} \emph{Remote Sensing of
Environment} 202: 18--27.

\leavevmode\vadjust pre{\hypertarget{ref-haibohe2009}{}}%
Haibo He, and E. A. Garcia. 2009. {``Learning from Imbalanced Data.''}
\emph{IEEE Transactions on Knowledge and Data Engineering} 21 (9):
1263--84. \url{https://doi.org/10.1109/tkde.2008.239}.

\leavevmode\vadjust pre{\hypertarget{ref-terra}{}}%
Hijmans, Robert J. 2021. \emph{Terra: Spatial Data Analysis}.
\url{https://CRAN.R-project.org/package=terra}.

\leavevmode\vadjust pre{\hypertarget{ref-Hobbs2009}{}}%
Hobbs, Richard J, Eric Higgs, and James A Harris. 2009. {``Novel
Ecosystems: Implications for Conservation and Restoration.''}
\emph{Trends in Ecology \& Evolution} 24 (11): 599--605.
\url{https://doi.org/10.1016/j.tree.2009.05.012}.

\leavevmode\vadjust pre{\hypertarget{ref-huang2019}{}}%
Huang, Jianfeng, Xinchang Zhang, Qinchuan Xin, Ying Sun, and Pengcheng
Zhang. 2019. {``Automatic Building Extraction from High-Resolution
Aerial Images and LiDAR Data Using Gated Residual Refinement Network.''}
\emph{ISPRS Journal of Photogrammetry and Remote Sensing} 151 (May):
91--105. \url{https://doi.org/10.1016/j.isprsjprs.2019.02.019}.

\leavevmode\vadjust pre{\hypertarget{ref-Johnson2006}{}}%
Johnson, Vanessa S, John A Litvaitis, Thomas D Lee, and Serita D Frey.
2006. {``The Role of Spatial and Temporal Scale in Colonization and
Spread of Invasive Shrubs in Early Successional Habitats.''}
\emph{Forest Ecology and Management} 228 (1-3): 124--34.
\url{https://doi.org/10.1016/j.foreco.2006.02.033}.

\leavevmode\vadjust pre{\hypertarget{ref-Kauth1976}{}}%
Kauth, Richard J., and G. S. P. Thomas. 1976. {``The Tasselled Cap - a
Graphic Description of the Spectral-Temporal Development of Agricultural
Crops as Seen by Landsat.''} In \emph{Symposium on Machine Processing of
Remotely Sensed Data}.

\leavevmode\vadjust pre{\hypertarget{ref-lightgbm}{}}%
Ke, Guolin, Qi Meng, Thomas Finley, Taifeng Wang, Wei Chen, Weidong Ma,
Qiwei Ye, and Tie-Yan Liu. 2017. {``LightGBM: A Highly Efficient
Gradient Boosting Decision Tree.''} In \emph{Advances in Neural
Information Processing Systems}, edited by I. Guyon, U. V. Luxburg, S.
Bengio, H. Wallach, R. Fergus, S. Vishwanathan, and R. Garnett. Vol. 30.
Curran Associates, Inc.
\url{https://proceedings.neurips.cc/paper/2017/file/6449f44a102fde848669bdd9eb6b76fa-Paper.pdf}.

\leavevmode\vadjust pre{\hypertarget{ref-Kennedy2010}{}}%
Kennedy, Robert E, Zhiqiang Yang, and Warren B. Cohen. 2010.
{``Detecting Trends in Forest Disturbance and Recovery Using Yearly
Landsat Time Series: 1. {LandTrendr} {\textemdash} Temporal Segmentation
Algorithms.''} \emph{Remote Sensing of Environment} 114 (12):
2897--2910. \url{https://doi.org/10.1016/j.rse.2010.07.008}.

\leavevmode\vadjust pre{\hypertarget{ref-Kennedy2018}{}}%
Kennedy, Robert E, Zhiqiang Yang, Noel Gorelick, Justin Braaten, Lucas
Cavalcante, Warren B. Cohen, and Sean Healey. 2018. {``Implementation of
the LandTrendr Algorithm on Google Earth Engine.''} \emph{Remote
Sensing} 10 (5). \url{https://doi.org/10.3390/rs10050691}.

\leavevmode\vadjust pre{\hypertarget{ref-King2014}{}}%
King, David I., and Scott Schlossberg. 2014. {``Synthesis of the
Conservation Value of the Early-Successional Stage in Forests of Eastern
North America.''} \emph{Forest Ecology and Management} 324: 186--95.
\url{https://doi.org/10.1016/j.foreco.2013.12.001}.

\leavevmode\vadjust pre{\hypertarget{ref-Kulmatiski2006}{}}%
Kulmatiski, Andrew, Karen H Beard, and John M Stark. 2006. {``Soil
History as a Primary Control on Plant Invasion in Abandoned Agricultural
Fields.''} \emph{Journal of Applied Ecology} 43 (5): 868--76.
\url{https://doi.org/10.1111/j.1365-2664.2006.01192.x}.

\leavevmode\vadjust pre{\hypertarget{ref-LeCun2015}{}}%
LeCun, Yann, Yoshua Bengio, and Geoffrey Hinton. 2015. {``Deep
Learning.''} \emph{Nature} 521: 436--44.
\url{https://doi.org/10.1038/nature14539}.

\leavevmode\vadjust pre{\hypertarget{ref-Lorimer2001}{}}%
Lorimer, Craig G. 2001. {``Historical and Ecological Roles of
Disturbance in Eastern North American Forests: 9,000 Years of Change.''}
\emph{Wildlife Society Bulletin (1973-2006)} 29 (2): 425--39.
\url{http://www.jstor.org/stable/3784167}.

\leavevmode\vadjust pre{\hypertarget{ref-terrainr}{}}%
Mahoney, Michael J., Colin M. Beier, and Aidan C. Ackerman. 2022.
{``{terrainr}: An {R} Package for Creating Immersive Virtual
Environments.''} \emph{Journal of Open Source Software} 7 (69): 4060.
\url{https://doi.org/10.21105/joss.04060}.

\leavevmode\vadjust pre{\hypertarget{ref-McCay2009}{}}%
McCay, Timothy S, and Deanna H McCay. 2009. {``Processes Regulating the
Invasion of European Buckthorn (Rhamnus Cathartica) in Three Habitats of
the Northeastern United States.''} \emph{Biological Invasions} 11 (8):
1835--44. \url{https://doi.org/10.1007/s10530-008-9362-7}.

\leavevmode\vadjust pre{\hypertarget{ref-NOAA}{}}%
NOAA National Centers for Environmental Information. 2022. {``Climate at
a Glance: Statewide Mapping.''} \url{https://www.ncdc.noaa.gov/cag/}.

\leavevmode\vadjust pre{\hypertarget{ref-Perring2013}{}}%
Perring, Michael P, Rachel J Standish, and Richard J Hobbs. 2013.
{``Incorporating Novelty and Novel Ecosystems into Restoration Planning
and Practice in the 21st Century.''} \emph{Ecological Processes} 2 (1):
1--8. \url{https://doi.org/10.1186/2192-1709-2-18}.

\leavevmode\vadjust pre{\hypertarget{ref-PRISM}{}}%
PRISM Climate Group. 2022. {``PRISM Climate Data.''}
\url{https://prism.oregonstate.edu}.

\leavevmode\vadjust pre{\hypertarget{ref-Python}{}}%
Python Core Team. 2022. \emph{{Python: A dynamic, open source
programming language}}. {Python Software Foundation}.
\url{https://www.python.org/}.

\leavevmode\vadjust pre{\hypertarget{ref-R}{}}%
R Core Team. 2021. \emph{R: A Language and Environment for Statistical
Computing}. Vienna, Austria: R Foundation for Statistical Computing.
\url{https://www.R-project.org/}.

\leavevmode\vadjust pre{\hypertarget{ref-lidR}{}}%
Roussel, Jean-Romain, David Auty, Nicholas C. Coops, Piotr Tompalski,
Tristan R. H. Goodbody, Andrew Sánchez Meador, Jean-François Bourdon,
Florian de Boissieu, and Alexis Achim. 2020. {``lidR: An r Package for
Analysis of Airborne Laser Scanning (ALS) Data.''} \emph{Remote Sensing
of Environment} 251: 112061.
\url{https://doi.org/10.1016/j.rse.2020.112061}.

\leavevmode\vadjust pre{\hypertarget{ref-Ruiz2018}{}}%
Ruiz, Luis Ángel, Jorge Abel Recio, Pablo Crespo-Peremarch, and Marta
Sapena. 2018. {``An Object-Based Approach for Mapping Forest Structural
Types Based on Low-Density LiDAR and Multispectral Imagery.''}
\emph{Geocarto International} 33 (5): 443--57.
\url{https://doi.org/10.1080/10106049.2016.1265595}.

\leavevmode\vadjust pre{\hypertarget{ref-Spiering2019}{}}%
Spiering, David J. 2019. {``Brownfields and Old-Fields: Vegetation
Succession in Post-Industrial Ecosystems of Western New York.''} PhD
thesis, State University of New York at Buffalo.

\leavevmode\vadjust pre{\hypertarget{ref-Stover1998}{}}%
Stover, Marian E, and PL Marks. 1998. {``Successional Vegetation on
Abandoned Cultivated and Pastured Land in Tompkins County, New York.''}
\emph{Journal of the Torrey Botanical Society}, 150--64.
\url{https://doi.org/10.2307/2997302}.

\leavevmode\vadjust pre{\hypertarget{ref-3DEP}{}}%
U.S. Geological Survey. 2019. {``{3D} Elevation Program 1-Meter
Resolution Digital Elevation Model.''}
\url{https://www.usgs.gov/core-science-systems/ngp/3dep/data-tools}.

\leavevmode\vadjust pre{\hypertarget{ref-NAIP}{}}%
US Department of Agriculture. 2019. {``National Agricultural Imagery
Program.''}

\leavevmode\vadjust pre{\hypertarget{ref-NASS}{}}%
USDA National Agricultural Statistics Service. 2019. {``2017 Census of
Agriculture.''} \url{www.nass.usda.gov/AgCensus}.

\leavevmode\vadjust pre{\hypertarget{ref-Whitney1994}{}}%
Whitney, Gordon G. 1994. \emph{From Coastal Wilderness to Fruited Plain:
A History of Environmental Change in Temperate North America from 1500
to the Present}. Cambridge, United Kingdom: Cambridge University Press.

\leavevmode\vadjust pre{\hypertarget{ref-Wickham2021}{}}%
Wickham, James, Stephen V. Stehman, Daniel G. Sorenson, Leila Gass, and
Jon A. Dewitz. 2021. {``Thematic Accuracy Assessment of the NLCD 2016
Land Cover for the Conterminous United States.''} \emph{Remote Sensing
of Environment} 257: 112357.
https://doi.org/\url{https://doi.org/10.1016/j.rse.2021.112357}.

\leavevmode\vadjust pre{\hypertarget{ref-Williams2007}{}}%
Williams, John W., and Stephen T. Jackson. 2007. {``Novel Climates,
No-Analog Communities, and Ecological Surprises.''} \emph{Frontiers in
Ecology and the Environment} 5 (9): 475--82.
\url{https://doi.org/10.1890/070037}.

\leavevmode\vadjust pre{\hypertarget{ref-Wolpert1992}{}}%
Wolpert, David H. 1992. {``Stacked Generalization.''} \emph{Neural
Networks} 5 (2): 241--59.
\url{https://doi.org/10.1016/S0893-6080(05)80023-1}.

\leavevmode\vadjust pre{\hypertarget{ref-ranger}{}}%
Wright, Marvin N., and Andreas Ziegler. 2017. {``{ranger}: A Fast
Implementation of Random Forests for High Dimensional Data in {C++} and
{R}.''} \emph{Journal of Statistical Software} 77 (1): 1--17.
\url{https://doi.org/10.18637/jss.v077.i01}.

\leavevmode\vadjust pre{\hypertarget{ref-NLCD}{}}%
Yang, Limin, Suming Jin, Patrick Danielson, Collin Homer, Leila Gass,
Stacie M. Bender, Adam Case, et al. 2018. {``A New Generation of the
United States National Land Cover Database: Requirements, Research
Priorities, Design, and Implementation Strategies.''} \emph{ISPRS
Journal of Photogrammetry and Remote Sensing} 146: 108--23.
\url{https://doi.org/10.1016/j.isprsjprs.2018.09.006}.

\leavevmode\vadjust pre{\hypertarget{ref-Youden1950}{}}%
Youden, W. J. 1950. {``Index for Rating Diagnostic Tests.''}
\emph{Cancer} 3 (1): 32--35.
\url{https://doi.org/10.1002/1097-0142(1950)3:1\%3C32::AID-CNCR2820030106\%3E3.0.CO;2-3}.

\leavevmode\vadjust pre{\hypertarget{ref-zarea2016}{}}%
Zarea, Asghar, and Ali Mohammadzadeh. 2016. {``A Novel Building and Tree
Detection Method from LiDAR Data and Aerial Images.''} \emph{IEEE
Journal of Selected Topics in Applied Earth Observations and Remote
Sensing} 9 (5): 1864--75.
\url{https://doi.org/10.1109/jstars.2015.2470547}.

\end{CSLReferences}

\end{document}